\documentclass[letterpaper]{article} 
\usepackage[]{aaai24}  
\usepackage{times}  
\usepackage{helvet}  
\usepackage{courier}  
\usepackage[hyphens]{url}  
\usepackage{graphicx} 
\usepackage{natbib}  
\usepackage{caption} 
\usepackage{algorithm}
\usepackage{algorithmic}

\usepackage{newfloat}
\usepackage{listings}
\DeclareCaptionStyle{ruled}{labelfont=normalfont,labelsep=colon,strut=off} 
\floatstyle{ruled}
\newfloat{listing}{tb}{lst}{}
\floatname{listing}{Listing}
\usepackage[cmyk]{xcolor}
\usepackage{amssymb}
\usepackage{array}
\usepackage{mathtools}
\usepackage{multirow}
\usepackage{subcaption}
\usepackage{tikz}

\definecolor{color1}{cmyk}{0.216,0.176,0,0}
\definecolor{color2}{cmyk}{0.059,0.235,0.392,0}

\title{Bootstrapping Cognitive Agents with a Large Language Model}
\author{
    Feiyu (Gavin) Zhu,
    Reid Simmons
}
\affiliations{
    Carnegie Mellon University

    feiyuz@andrew.cmu.edu, rsimmons@andrew.cmu.edu
}

\usepackage{bibentry}

\begin{document}

\maketitle

\begin{abstract}

Large language models contain noisy general knowledge of the world, yet are hard to train or fine-tune. On the other hand cognitive architectures have excellent interpretability and are flexible to update but require a lot of manual work to instantiate. In this work, we combine the best of both worlds: bootstrapping a cognitive-based model with the noisy knowledge encoded in large language models. Through an embodied agent doing kitchen tasks, we show that our proposed framework yields better efficiency compared to an agent based entirely on large language models. Our experiments indicate that large language models are a good source of information for cognitive architectures, and the cognitive architecture in turn can verify and update the knowledge of large language models to a specific domain.

\end{abstract}


\section{Introduction}

Large language models (LLM) such as GPT-4 \cite{openai2023gpt4}, have shown emerging capabilities after training on internet-scale text data with human feedback, and have been employed in robot planning \cite{huang2022language}, animal behavior analysis \cite{ye2023amadeusgpt},  human proxies \cite{zhang2023large}, and many more.
However, they have also been criticized for being susceptible to adversarial attacks \cite{zou2023universal}, hallucination \cite{casper2023open}, and having diminishing returns for scaling \cite{openai2023gpt4}.

Cognitive architectures are another approach in the pursuit of artificial general intelligence that attempt to unify all aspects of human cognition computationally \cite{newell1994unified}.
Despite the variety of architectures developed, most of them share the same central components, consisting of declarative memory reflecting knowledge of the world, procedural memory dictating the agent's behavior given certain scenarios, and short-term working memory that assists reasoning and planning \cite{laird2017standard}.

The procedural memory is represented by a set of production rules, each with a precondition and an effect.
Agents operate in perceive-plan-act cycles, dynamically matching relevant features of the environment to the production rules and applying their effects.
Unlike operators in symbolic planning, production rules do not represent alternative actions, but instead reflect different contextual knowledge \cite{laird2022introduction}.
These rules can be reinforced and modified throughout the agent's learning process.
Despite some pioneering work on data-driven cognitive model creation \cite{hake2022inferring}, almost all previous work generate their initial set of production rules manually, limiting their application to simple environments such as blocks world or psychology experiments \cite{park2023generative}.

\begin{figure}
    \centering
    \resizebox{0.98\columnwidth}{!}{%
    \begin{tikzpicture}

    \fill[gray!20!white, rounded corners=16pt] (1,-6.5) rectangle (17,7.5);
    
    \fill[white, rounded corners=16pt] (1.5, -3.3) rectangle (16.5, 0);
    \fill[white, rounded corners=16pt] (1.5, -6) rectangle (16.5, -4);
    \fill[white, rounded corners=16pt] (9.5, 1) rectangle (16.5, 5);
    \fill[white, rounded corners=16pt] (1.5, 1) rectangle (8.5, 5);

    \node[inner sep=0pt] (production_slice) at (4, -1.8) {\includegraphics[width=0.25\columnwidth]{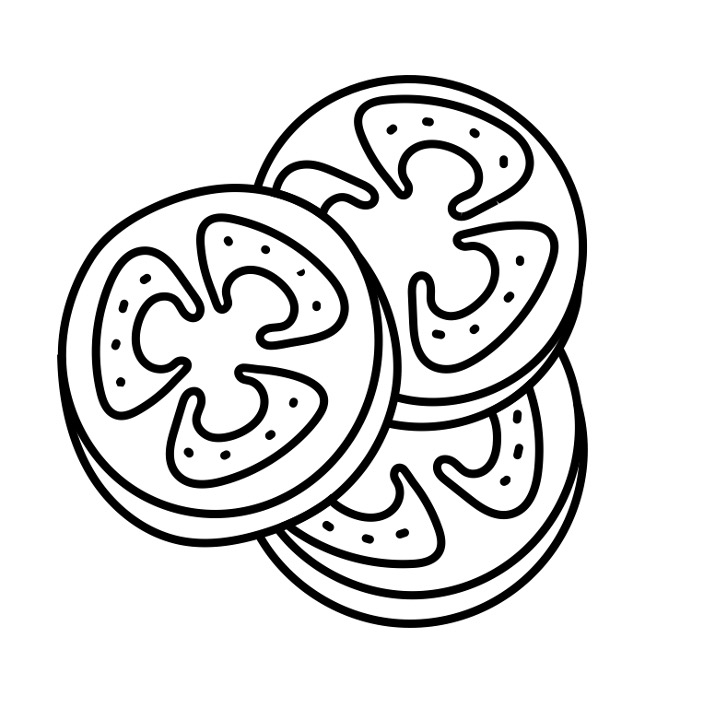}};
    \node[inner sep=0pt] (production_gripper) at (6.5, -1.8) {\includegraphics[width=0.25\columnwidth]{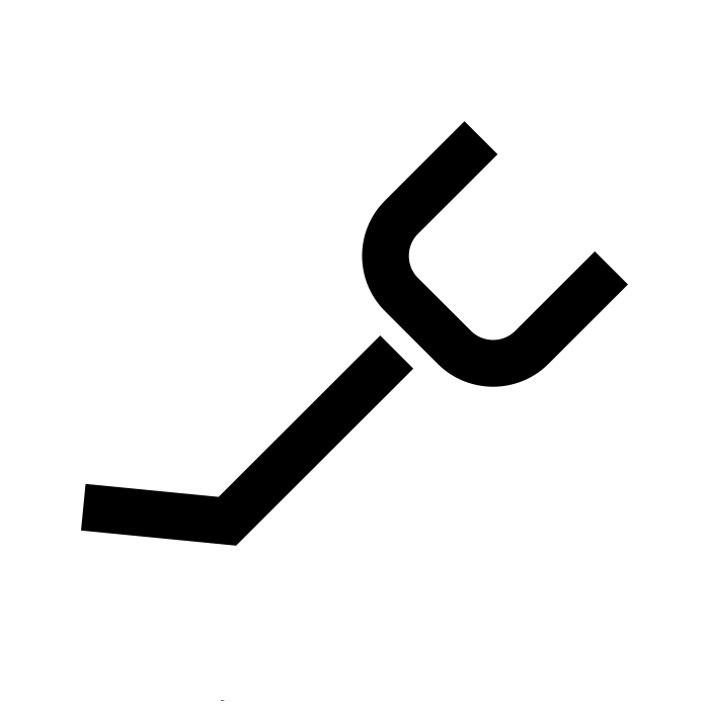}};
    \node[inner sep=0pt] (production_table) at (10.5, -1.8) {\includegraphics[width=0.25\columnwidth]{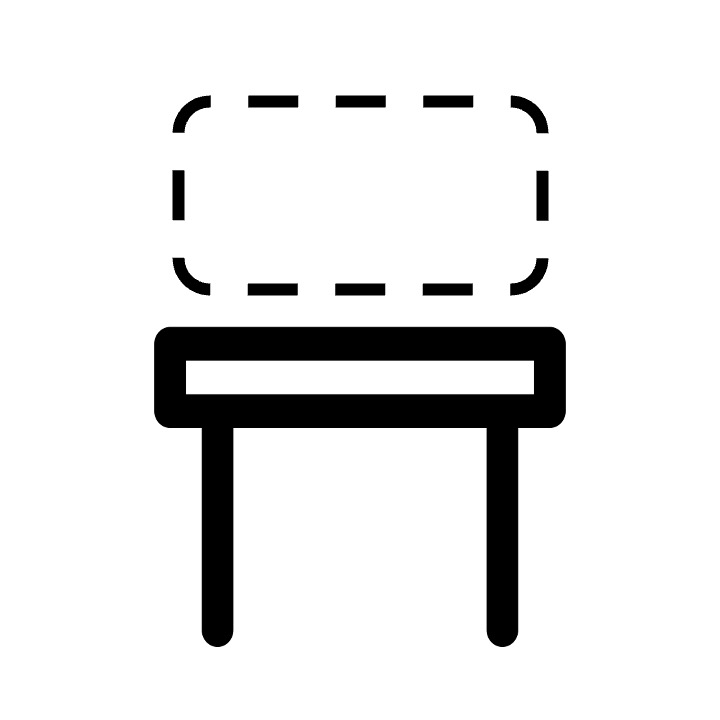}};
    \node[inner sep=0pt] (production_find) at (14, -1.5) {\includegraphics[width=0.25\columnwidth]{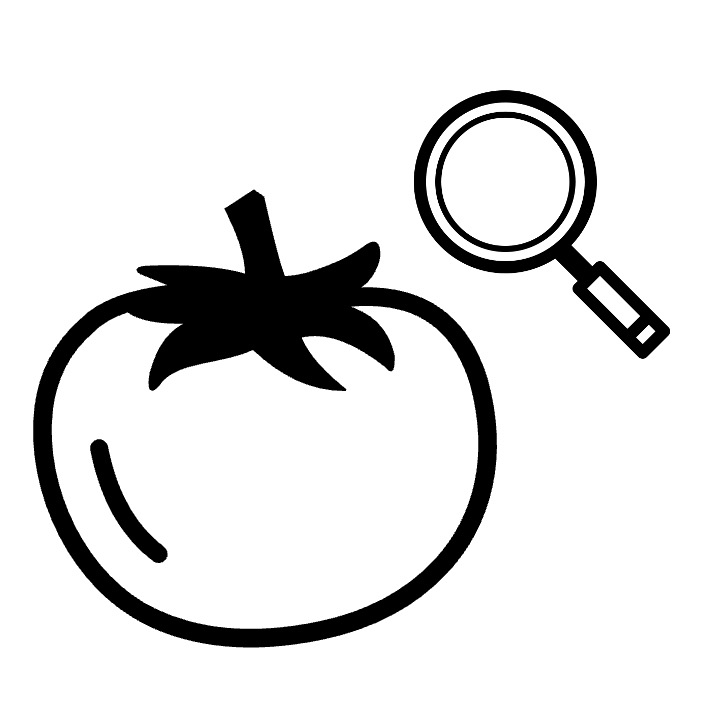}};
    \node[scale=3] at (5, -1.8) { $ \colon$ };
    \node[scale=2] at (8.5, -1.8) { AND };
    \node[scale=3] at (12.1, -1.9) { $ \rightarrow$ };
    
    \node[inner sep=0pt] (world_tomato) at (3.5, 2.1) {\includegraphics[width=0.25\columnwidth]{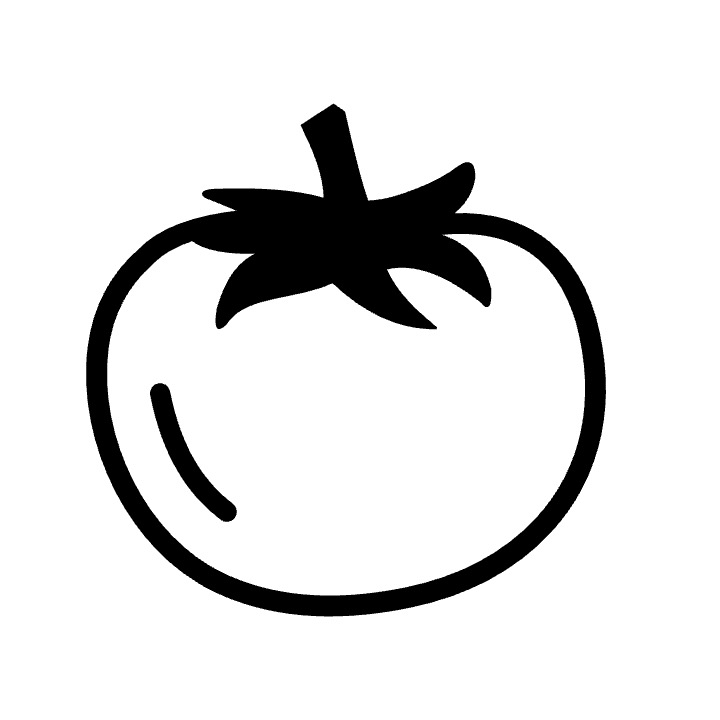}};
    \node[inner sep=0pt] (world_fridge) at (6.5, 2.1) {\includegraphics[width=0.25\columnwidth]{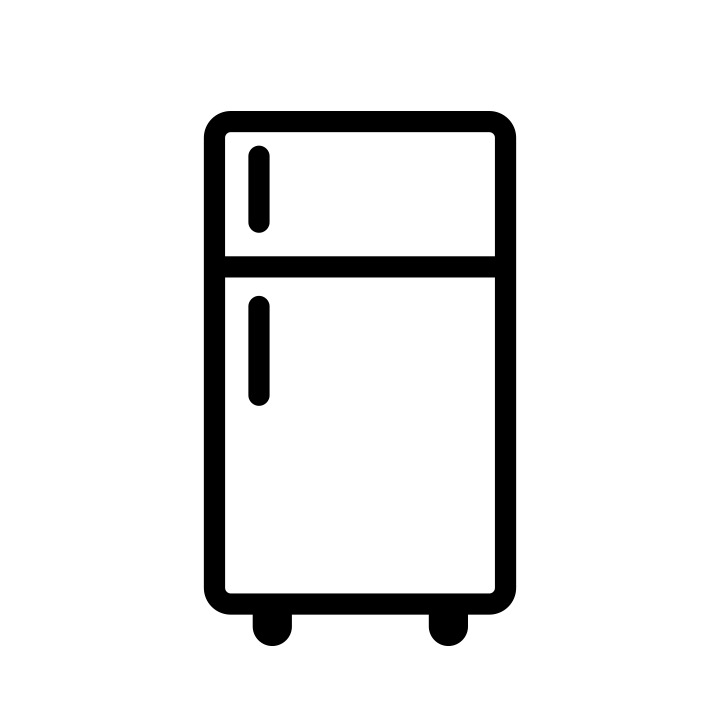}};
    \node[scale=3] at (5.1, 2) { $ \Rightarrow$ };

    \node[inner sep=0pt] (env_gripper) at (11.5, 2.1) {\includegraphics[width=0.25\columnwidth]{figures/icons/empty_gripper3.jpeg}};
    \node[inner sep=0pt] (env_table) at (14.5, 2.1) {\includegraphics[width=0.25\columnwidth]{figures/icons/empty_table.jpeg}};
    
    \node[inner sep=8pt] (env) at (18.7, 0.5) {\includegraphics[width=0.3\columnwidth]{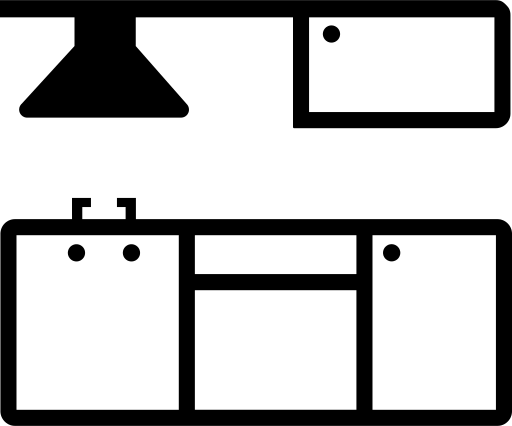}};
    \node[inner sep=0pt] (robot) at (9, 6.3) {\includegraphics[width=0.3\columnwidth]{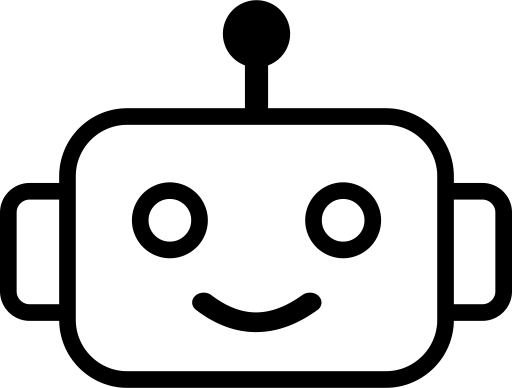}};
    \node[inner sep=8pt] (openai) at (9, 9) {\includegraphics[width=0.2\columnwidth]{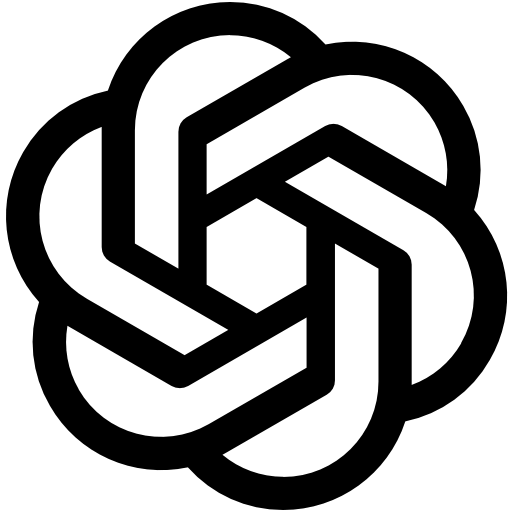}};
    \node[scale=2.8] at (9, 10.5) {LLM};
    
    \node[scale=2.2, text width=0.78\columnwidth] at (9, 4.5) { World Knowledge: };
    \node[scale=1.5] at (5, 3.6) { (tomatoes belong in the fridge) };
    \node[scale=2.2, text width=0.35\columnwidth] at (13, 4.5) { Envt Knowledge: };
    \node[scale=1.5] at (13, 3.6) { (gripper empty \& table empty) };
    \node[scale=2.2, text width=0.78\columnwidth] at (9, -0.5) { Productions: };
    \node[scale=1.5] at (4, -3) { (goal) };
    \node[scale=1.5] at (13.8, -3) { (subtask: find a tomato) };
    \node[scale=2.2, text width=0.78\columnwidth] at (9, -5) {
        \hspace{1in }Task Stack: \\
        \texttt{slice a tomato} $~~~~~~~$ \texttt{find a tomato}
    };
    \node[scale=2,] at (15, -3.8) { push };
    \node[scale=2,] at (13.5, 9.5) { query };
    \node[scale=2,] at (4, 9.5) { generate };
    \node[scale=2,] at (19, 3.5) { observe };
    \node[scale=2,] at (19, -2.3) { act };

    \draw[-latex, black, line width=4pt, rounded corners=16pt] (openai.west) -- (0.1, 9) -- (0.1, -1.5) -- (1.5, -1.5);
    \draw[-latex, black, line width=4pt, rounded corners=16pt] (13, 5) -- (13, 9) -- (openai.east);
    \draw[-latex, black, line width=4pt, rounded corners=16pt] (0.1, 2.5) -- (1.5, 2.5);
    \draw[-latex, black, line width=4pt, rounded corners=16pt] (env.north) -- (18.7, 3) -- (16.5, 3);
    \draw[latex-, black, line width=4pt, rounded corners=16pt] (env.south) -- (18.7, -2) -- (16.5, -2);
    \draw[-latex, black, line width=4pt, rounded corners=16pt] (13.8, -3.3) -- (13.8, -4.8);
    
    \draw[line width=4pt, rounded corners=16pt, dashed] (5, 1) -- (5, 0.3) -- (13, 0.3) -- (13, 1);
    \draw[-latex, line width=4pt, rounded corners=16pt, dashed] (9, 0.3) -- (9, -1);
    \draw[-latex, line width=4pt, rounded corners=16pt, dashed] (4, -4.8) -- (4, -3.4);


    
    \end{tikzpicture}
    }
    \caption{Overview of agent framework. It is showing the agent executing the production of attending to a new subtask of finding a  tomato when the original task is to slice a tomato and the tomato is not in the gripper nor on the table. Dotted lines represent the information a production rule may condition on. Solid lines represent information flow.}
    \label{fig:overview}
\end{figure}

In this work we combine the two approaches in a complementary fashion (Figure \ref{fig:overview}).
LLMs encode the common sense knowledge of the world \cite{madaan2022language} that can be used in place of human labor for constructing agents in the cognitive architecture.
And the reasoning and learning capabilities in the cognitive architecture can identify and filter the noise in LLMs, while converting the knowledge in language to actionable productions of an embodied agent.

This combined framework separates knowledge generation and knowledge application, and this modularity is the key to generalization.
The LLM is responsible only for generating general knowledge, such as ``if the task is to find an object, the agent should explore the places where that object is commonly stored''.
Since such knowledge can be applied to almost all objects and environments, the LLM needs to generate these only once, and it is the role of the cognitive architecture to dynamically match the environment to the generated knowledge.
This is significantly different from using LLMs to generate plans directly, as the plans are grounded to the specific instance of the task (e.g., finding a specific object in the specific environment), and are non-trivial to generalize to novel environments without re-generation.

The contribution of this paper is threefold: 1) we propose an agent framework that combines LLMs with customized cognitive architecture, 2) we demonstrate how it can learn to perform various kitchen tasks from bootstrapping, and 3) we show that, when applied to new environments, it requires significantly fewer tokens than querying LLM for actions.


\section{Related Work}

\subsection{Learning Through Program Synthesis}

Interactive Task Learning (ITL) \cite{laird2017interactive} aims at teaching robots new skills in a one-shot fashion.
Previous work implements this in the SOAR cognitive architecture and has shown effective task and environment transferability in domains such as board games \cite{kirk2019learning} and embodied agents \cite{mininger2022demonstration}.
To reduce the need for extensive human input, recent research explores using LLM as the knowledge source \cite{lindes2023improving, kirk2023integrating}, shifting human labor from specifying the goal conditions to answering yes/no questions.
In contrast, our approach uses strategic prompting and self-reflection mechanisms to eliminate the need for human supervision.



Our work shares some high-level ideas with DreamCoder \cite{ellis2021dreamcoder}, which learns to solve new problems by program generation and reflection.
Instead of formulating it as an informed search problem, we accelerate this process by querying LLMs for their existing knowledge.

\citeauthor{madaan2022language} (\citeyear{madaan2022language}) extract common sense knowledge from LLMs into code form similar to how we extract productions.
But they only address the general task decomposition, not applying the information to an embodied agent.

\subsection{Large Language Model for Embodied Agents}

Many studies have explored using LLMs to generate code that performs robotics tasks \cite{liang2023code, singh:progprompt, vemprala2023chatgpt} and game environments \cite{wang2023voyager}, which is similar to the procedural memory in the cognitive architectures.
In addition to script-based code, other works explored generating PDDL specifications \cite{liu2023llm+, xie2023translating}.
Unlike the situation-grounded code produced by these methods, our approach generates parameterized productions with learnable weights.
This allows more generalization capabilities and choosing the best plan among multiple applicable plans.

Others let LLMs select the action directly \cite{di2023towards, vemprala2023chatgpt} with the help of other auxiliary components such as affordance evaluation \cite{ahn2022can}, memory stream \cite{park2023generative}, visual summarization \cite{qiu2023embodied}, and knowledge base \cite{zhu2023ghost}.
Some others explored multi-modal foundation models tailored for embodied agents \cite{driess2023palm, xiang2023language}.
As LLMs are non-trivial to update from a single instance, using more explicit production systems in our approach enables persistent one-shot updates and more interpretability.
As we will show in our experiments, relying on LLMs for every action is also not very cost-effective.


\section{Method}

\subsection{Architecture Overview}

Figure \ref{fig:overview} illustrates the architecture and workflow of the agent.
The agent has four main components.
A world knowledge base that contains general knowledge, such as ``Tomatoes are commonly stored in the Fridge''.
An environment knowledge that reflects what the agent knows about the environment from past observations, including both information about the agent itself (e.g., the gripper is empty) and about the external world (e.g., the table is clear).
These two components form the declarative memories of the agent.

Another essential component is the procedural memory that contains all the production rules.
In our work, however, we integrate the working memory into each production by exploiting the Python class structure, so there is no centralized working memory.
And finally inspired by the goal module of ACT-R \cite{anderson2009can} and the impasse mechanism of SOAR \cite{laird2022introduction}, the agent manages a stack of tasks.

At each time step, the agent searches in its procedural memory for any applicable production rule, considering the current task and environment knowledge.
If there is no production applicable, the agent will summarize the current knowledge and query an LLM for both an action suggestion, and a corresponding production rule, such that the agent knows what to do in similar scenarios in the future.
When at least one production applicable, it will sample an applicable production rule, based on its utility, and execute the proposed action, which can be either in the environment or internally, such as adding a subtask to its task stack.

\subsection{Bootstrapping Procedures}

The bootstrapping process starts with a \textit{curriculum}. 
We took inspiration from \cite{wang2023voyager}, which uses an LLM to automatically construct the curriculum for the game of Minecraft.
As the simulator we are using is not as popular as Minecraft, and the robot has some very specific affordance model (e.g., can only hold one object at a time), we find it better to specify the curriculum manually.

Another difference is that our curriculum consists of families of tasks (e.g., \texttt{find a/an <object>}) instead of specific instances (e.g., \texttt{find a/an egg}).
We follow the SOAR syntax and keep all variables in angle brackets.

Unlike previous works that require human input on the next steps and/or goal condition for the tasks \cite{mininger2022demonstration}, we only require the names of the task families, so designing the curriculum is not very labor intensive.

With a given curriculum, the following steps are used to bootstrap a single task in the curriculum (using \texttt{find a/an <object>} as an example).
\begin{enumerate}
    \item Fill in the variables randomly from the environment to instantiate a concrete task (e.g., \texttt{find a/an Egg});
    \item Attempt the task with the existing production rules;
    \item (\textbf{Action Selection}) If there is no production rule for a state, or there is a cycle detected through the production application, query an LLM for an action;
    \item (\textbf{Production Generation}) Generate the corresponding production rule to the action, and load it into the agent;
    \item Repeat step 1-4 sufficient times until the robot can perform the task with only production rules;
    \item (\textbf{Production Improvement}) Use a critic to summarize the end condition of the task for future use and improve the generated productions.
\end{enumerate}

The above procedures are repeated for all task families in the curriculum.
While the agent might not fully learn every scenario of a task before moving on to the next one, it can still query LLM to generate a production rule for a previous task later. 
The training of a task is considered complete as long as it has sufficient experience with the task to generate a reasonable end condition such that future tasks can reuse the previously learned tasks.

\subsection{Action Selection}\label{action selection}

The LLM is prompted with the current task, a summary of the current state, and a list of options available to the robot, which include both motor actions on the environment (e.g., move to a specific location) and internal action (e.g., attend to a new subtask).
For each previously trained subtask, we provide the end condition generated by the critic for the LLM to evaluate its relevance.
Like the task names, the actions can also be parameterized (e.g., \texttt{move to <receptacle>}) and the LLM can replace \texttt{<receptacle>} with anything as it sees fit.

We use chain-of-thought prompting \cite{wei2022chain}, which explicitly instructs the LLM to respond to the prompt in a step-by-step manner, probing it to make the most informed decision.
The LLM is instructed to reflect on common strategies for approaching the task, analyze the current situation, and evaluate the usefulness of each action before suggesting one option for the robot to take.
The LLM is also prompted to state the purpose of the chosen action, which will inform the production rule generation later.

\subsection{Production Generation}

Although the production rule is generated based on the current state, they are not plans for the current task but instead should be the underlying decision-making principle for all similar scenarios.
For example, if the current task is to \texttt{find a/an egg}, instead of suggesting the action sequence of exploring every cabinet in the current environment, a desirable production rule would suggest ``whenever you need to find something, you should first explore the unexplored places where that object is commonly stored''.
This is a systematic generalization that can be applied to finding any objects, not just eggs or food, and also to other novel environments with different layouts and receptacle types.

To generate desired production rules, we separate the production generation into two steps. The first step summarizes the action selection process and generates the English description of the production rule; the second step converts the description into executable Python code (Listing \ref{lst:production}).
This separation is inspired by how human beginners are instructed to build cognitive models \cite{laird2017tutorial}, and has two benefits.
First, it allows each query to the LLM to be of reasonable length ($\sim 5k$ tokens), preventing LLMs to lose focus on overly long prompts \cite{liu2023lost}.
Second, it enables a modular design, which allows generating code from English descriptions generated from other sources, including human feedback and post-generation self-reflection.

For each step, we also use the chain-of-thought prompting technique.
For English description generation, the LLM is given the entire history of the action selection process, and is instructed to take four steps: 1) identify relevant information that leads to choosing the action, 2) generate a specific production rule that describes the current situation, 3) identify the potentially generalizable components in the specific rule and what they can be generalized to, and 4) replace the components to form the generalized production description.

For code generation, the LLM is given the Python interface of querying declarative memory and the current task, and is instructed to take another four steps: 1) plan what variable bindings are needed, and how their values should be assigned, 2) analyze the predicates in the precondition and associate them with relevant variables, 3) plan how each predicate should be tested using the provided function interfaces, and 4) fill in the production template.

The code snippet is parsed from the LLM output, saved as a Python file, and dynamically imported into the agent.

\begin{listing}[tb]%
\caption{Production interface}%
\label{lst:production}%
\begin{lstlisting}[language=Python]
class GeneratedProduction(Production):
  def precondition(self, agent) -> bool:
    # Returns whether the production is applicable given the agent
    # Set variables as side-effects
  def apply(self) -> str:
    # Returns the effect
    # Based on the variable bindings
\end{lstlisting}
\end{listing}


\subsection{Production Improvement}

We use three mechanisms to monitor and improve the common interface mismatch, over-constraining, and over-generalization problems of the LLM-generated productions.

Similar to the iterative prompting design in Voyager \cite{wang2023voyager}, we replay the generated production rule on the state that it is generated, and ensure that its precondition check passes the existing declarative knowledge.
This mostly fixes errors regarding function interfaces, as the generated production has to comply with a specific naming scheme and the interface of the declarative knowledge.

Passing the precondition test for a single instance does not guarantee that production is ideal.
As the LLM has access to accumulated observations from the past during the action selection process, it might include unnecessary conditions that happen to be true in the precondition of the production, making it over-constraining.
This is handled by a critic LLM that summarizes the end condition of the task and provides suggestions on the existing productions.

Specifically, the critic LLM is given the name of the task family (e.g., \texttt{find a/an <object>}), and the English descriptions of the existing production rules for that task.
The LLM is instructed to first analyze all the production rules whose effect is the \texttt{done} action, and summarize the end condition of the given task in a sentence (e.g., \texttt{the robot is holding the desired object in its gripper}).
These end conditions summarize the behavior of the previously learned tasks to inform the action selection process for future tasks.
As mentioned in the action selection section, this summary will be added to the prompt when querying for tasks later in the curriculum to incentivize reusing previously learned skills.
Next, the LLM will go through all the production rules, and suggest modifications for each of them.
The LLM is also given the choice of keeping a production rule as is or removing it entirely.
The modifications are in the English description space for the critic, and we make use of the two-step modularity of production generation to update the production rules.



Over-generalization happens when important features are left out of the production's precondition.
For example, for the \texttt{pick and place} task, the LLM might generate a production rule that says:
\begin{verbatim}
 IF task is pick and place <object> AND
    <object> in field of view AND
    gripper is empty
 THEN pick <object>
\end{verbatim}
This will make the robot pick up the object even when the object is already in the target receptacle.
To prevent the agent from being stuck in an infinite loop, it will keep a state transition graph during the execution process and query the LLM for an alternative action once a cycle is detected using a depth-first search on the transition graph.
Coupled with the production reinforcement (described below), the agent will prioritize loop-breaking productions.

\subsection{Production Reinforcement}

Following previous work in visual navigation \cite{anderson2018evaluation}, the agent has to explicitly choose the special \texttt{done} action to indicate that it has completed the current task.
We further extend this and give the agent a \texttt{quit} option to indicate that it believes the given task is impossible for the given environment.
This is important as we allow the architecture to choose to attend to any subtask as it wants, and it should be able to realize when a task is impossible.

%

As we do not pre-define the goal condition during the bootstrapping process, we give a unit reward whenever the agent decides it is done with the current task.
The reward propagates back through the shortest path to the starting state.
For example, if the state transition is

\[ S_0 \xRightarrow[]{P_1} S_1 \xRightarrow[]{P_2} S_2 \xRightarrow[]{P_3} S_0 \xRightarrow[]{P_4} S_4 \xRightarrow[]{P_5} S_5 \xRightarrow[]{P_\text{done}} \]

where $S_0$ is the start state and $P_\text{done}$ is the production that yields the \texttt{done} action. Then the shortest path is

\[ S_0 \xRightarrow[]{P_4} S_4 \xRightarrow[]{P_5} S_5 \xRightarrow[]{P_\text{done}} \]


Therefore only $P_4, P_5, P_\text{done}$ will receive a utility update, using the bellman backup \cite{sutton2018reinforcement}.
\begin{align}
U_\mathrm{after}(P) &\gets \dfrac{1}{N(P) + 1} \left( N(P) \cdot U_\mathrm{before}(P) + \gamma^{\Delta_t} \right)
\end{align}
Where $U(P)$ is the utility of production $P$, $N(P)$ is the number of times $P$ gets applied, $\Delta_t$ is the time difference from production application to the \texttt{done} action, and $\gamma$ is the discount factor (which is set to $0.95$ for our experiments).

When a subtask is involved, the utility is updated with respect to each task. For example, if the state transition is

\[ A_0 \xRightarrow[]{P_1} A_1 \xRightarrow[]{P_2} \underbrace{B_3 \xRightarrow[]{Q_3} B_4 \xRightarrow[]{Q_4} B_5 \xRightarrow[]{Q_\text{done}}}_{\textrm{a subtask initiated by }P_2} A_6 \xRightarrow[]{P_\text{done}}\]

Where $A$ and $P$ correspond to the states and productions of the original task respectively and $B$ and $Q$ correspond to the states and productions of the subtask respectively.
This will be treated as two separate utility update pathways
\begin{align*}
  &A_0 \xRightarrow[]{P_1} A_1 \xRightarrow[]{P_2} A_6 \xRightarrow[]{P_\text{done}} \\
  &B_3 \xRightarrow[]{Q_3} B_4 \xRightarrow[]{Q_4} B_5 \xRightarrow[]{Q_\text{done}}
\end{align*}

If a subtask ends up with \texttt{quit} then there will be no utility update, not even negative ones.
Because the task might be impossible due to environmental constraints, which has nothing to do with the production rules.

Intuitively the closer a production brings the agent to choose \texttt{done} for its current task, the higher its utility will be.
This process is not provided to the LLM, so it has no incentive to ``cheat" by proposing the \texttt{done} action all the time.
We also explicitly tell the LLM to avoid selecting \texttt{done} or \texttt{quit} action unless it is ``absolutely certain'' about it.
This works empirically in our experiments.

This utility update process helps reduce the impact of hallucination in LLMs as the knowledge is aggregated.
For example, when tasked with ``explore the countertops'', the LLM may hallucinate and propose a production $P_\text{bad}$ that keeps the agent exploring the cabinets after all countertops have been explored, instead of proposing the \texttt{done} action, as it should.
However, when tasked with ``explore the sink`` in the same bootstrapping section, the LLM may generate a production $P_\text{good}$ that correctly identifies the termination condition and proposes \texttt{done} when all receptacles of the desired type have been explored.
Then later, when the agent needs to explore all the countertops (potentially as a subtask of another task) and all of the countertops have been explored, both $P_\text{bad}$ and $P_\text{good}$ will be applicable.
The agent will prioritize $P_\text{good}$ because it is guaranteed to have a higher utility value than $P_\text{bad}$.
On the other hand, if we use LLM to generate plans for each task, we would get a correct plan for the sink but an incorrect one for the countertops.

When multiple productions are applicable given the same environment knowledge, we resolve the conflict using the definition of noisy-optimal in previous works \cite{tian2023towards}. Where the probability of production $P_i$ being selected and applied, given the current knowledge $\mathcal{K}$, is
\begin{align}
\mathbb{P}(P_i \mid \mathcal{K}) &\propto \mathbf{I}_{\mathcal{K}}(P_i) \cdot \exp(U(P_i))
\end{align}
where $\mathbf{I}_\mathcal{K}(p)$ indicates that the preconditions of production $p$ hold, given knowledge $\mathcal{K}$.

\subsection{World Knowledge Base}

For the sake of simplicity, we implemented the world knowledge of the agent as a dictionary that maps natural language statements to either true or false.
Unlike many existing cognitive architectures that assume an absence of knowledge means the negation is true, we explicitly differentiate between not knowing and knowing to be false.

When a production rule is conditioned on a statement not previously known to the agent, the LLM is used to evaluate whether the statement is true, and the result will be saved to the knowledge base to be reused later.
For instance, when bootstrapping the task of finding an egg, the agent will learn the production rule that says ``If there is an unexplored receptacle where the object is commonly stored, explore that receptacle''.
But the agent does not know whether ``egg is commonly stored in the fridge'' is true or not initially, so it will query the LLM and memorize the positive response in its world knowledge base.
Later when the agent is tasked to put things in their common storage place, the agent can reuse the knowledge and place eggs into the fridge.
In addition to transferring to new tasks, the knowledge can be applied to new environments as well (e.g., eggs are commonly stored in fridges in most American households).

This knowledge base could be easily replaced by connecting it to an existing knowledge graph or ontology, but for the purpose of this paper, we are bootstrapping it from scratch.


\begin{figure}[ht]
    \centering
    \begin{subfigure}[b]{0.22\textwidth}
        \centering
        \includegraphics[width=\textwidth]{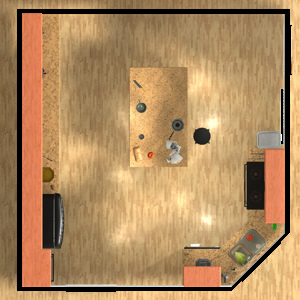}
        \caption{training floor plan}
        \label{fig:train}
    \end{subfigure}
    \hfill
    \begin{subfigure}[b]{0.22\textwidth}
        \centering
        \includegraphics[width=\textwidth]{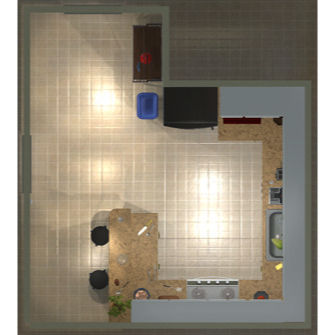}
        \caption{testing floor plan}
        \label{fig:test}
    \end{subfigure}
    \\
    \begin{subfigure}[b]{0.22\textwidth}
        \centering
        \includegraphics[width=\textwidth]{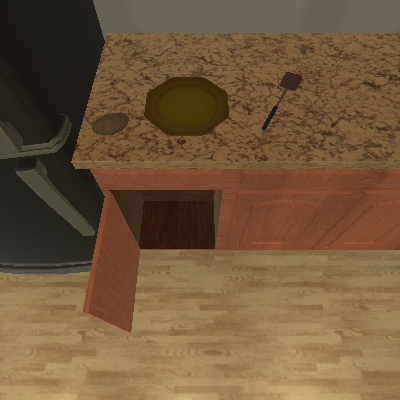}
        \caption{ego-centric view}
        \label{fig:ego}
    \end{subfigure}
    \hfill
    \begin{subfigure}[b]{0.22\textwidth}
        \centering
        \includegraphics[width=\textwidth]{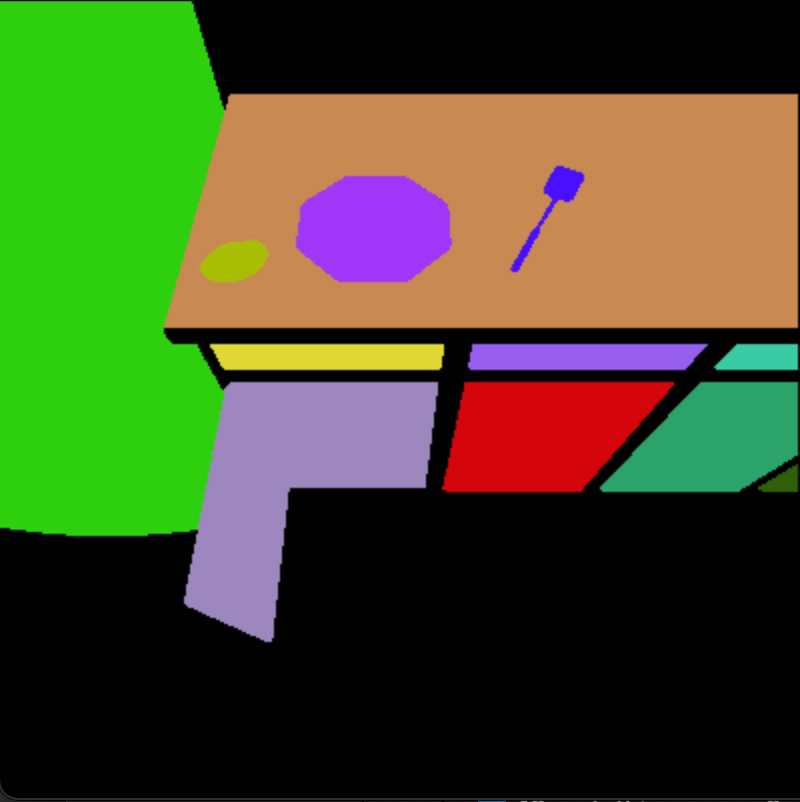}
        \caption{instance segmentation}
        \label{fig:instance}
    \end{subfigure}
    \caption{Screenshots of the AI2THOR simulator}
    \label{fig:ai2thor}
\end{figure}

\section{Experiments}

\subsection{Setup}

Following previous works in the embodied agents domain \cite{sarch2022tidee, trabucco2023a}, we evaluate our method in the kitchen environments in the AI2THOR simulator \cite{ai2thor}, shown in Figure \ref{fig:ai2thor}.
As shown in Figure \ref{fig:instance}, the agent has access to classification labels and attributes (e.g., ``is opened") for objects that are close enough (within $1.6m$) or large enough (more than $5\%$ of the frame).
We also assume the agent already knows the names and locations of the large receptacles (e.g., cabinets, fridges, etc.) but does not know what objects are in the receptacles until the agent actively explores them.

We use three different tasks for evaluation:
\begin{itemize}
    \item \texttt{find a/an <object>}: the goal is to have the specified object in the robot's field of view. 
    This is a fundamental skill that is often overlooked or directly assumed in many of the previous works \cite{singh:progprompt}.
    We want to show that our framework can bootstrap very basic skills in addition to composite actions.
    \item \texttt{slice a/an <object>}: the goal is to use a knife to slice an object.
    Because the robot can hold at most one item at a time, slicing involves a sequence of actions including finding the target object and the knife, putting them in the same place, and the final slice action.
    We want to show that our framework can handle tasks that involve multiple steps and tool use.
    \item \texttt{clear the countertops}: the goal is to have all the objects on the countertops moved to suitable storage places.
    This is a common household task that is investigated a lot in previous work \cite{habitatrearrangechallenge2022, sarch2022tidee}.
    We want to show that our framework can handle tasks that involve repeating similar subtasks.
\end{itemize}
The goal conditions listed above are only used for evaluation purposes, but are not provided to the LLM during training or testing.
The LLM has to infer the goal condition from the task description only.

For \texttt{find} and \texttt{slice}, $5$ target objects are chosen for each task, and we run $3$ trials for each object where the initial locations of the objects are shuffled.
For \texttt{clear the countertops} we run $3$ trials each with $5$ objects on the countertops that need to be put away.
The specific objects and locations vary between trials, and the success of the agent is evaluated based on how many objects originally on the countertops have been relocated to other places.
This results in $15$ specific goal instances for each task family.

We use \texttt{GPT4-0613} \cite{openai2023gpt4} for our experiments as previous works have shown that GPT3.5 is insufficient for code generation \cite{olausson2023demystifying, wang2023voyager}.
We set temperature to the $0$ for the most deterministic response.

\subsection{Conditions}

For the experiment condition, we bootstrapped our agent with the following curriculum in the training floor plan:
\begin{verbatim}
 1. explore <receptacle>
 2. find a/an <object>
 3. pick and place a/an <object> 
        in/on a/an <receptacle>
 4. slice a/an <object>
 5. put things on the countertop away
\end{verbatim}
This process generated $27$ production rules in total.
During test time, the agent can query the LLM for an immediate action if it does not have an applicable production rule for the current situation, but it cannot learn new production rules.

For the baseline condition of using LLMs to query only the actions, we omit the production generation steps and only use the action selection process within our framework.
This ensures the prompts used by both conditions are the same, so LLM should suggest actions of similar quality.
If the action proposed by the LLM leads to an affordance error, we query LLM another two times, and if none of the actions are viable by the agent, then it raises a failure.

Although many works address the rearrangement task \cite{sarch2022tidee, wu2023tidybot}, they are not appropriate baselines as their architectures already encode the general strategies (e.g., first determine the target receptacle for each object, then navigate to the target area, etc.) while our approach bootstraps everything from scratch.
Other code generation works cannot handle multiple instances of the same kind \cite{singh:progprompt} or understand the slicing preconditions \cite{song2022llm} without non-trivial modifications.

\subsection{Results}

\begin{table*}[ht]
\centering
\begin{tabular}{clcccc}
    Task & Agent & Success $\uparrow$ & Success w/o LLM $\uparrow$ & Steps $\downarrow$ & Tokens $\downarrow$ \\ \hline
    \multirow{ 2}{*}{\texttt{find a/an <object>}} & action-only & $14 / 15$ & - & $15.67$ & $54754.20$ \\ 
    & bootstrapped (ours) & $15 / 15$ & $12 / 15$ & $15.80$ & $916.87$\\ \hline
    \multirow{ 2}{*}{\texttt{slice a/an <object>}} & action-only & $15 / 15$ & - & $28.20$ & $102806.60$ \\ 
    & bootstrapped (ours) & $15 / 15$ & $15 / 15$ & $29.13$ & $0.00$ \\ \hline
    \multirow{ 2}{*}{\texttt{clear the countertops}} & action-only & $15 / 15$ & - & $5.13$ & $18924.87$ \\ 
    & bootstrapped (ours) & $15 / 15$ & $15 / 15$ & $7.47$ & $0.00$ \\ \hline
\end{tabular}
\caption{Result of experiments on household tasks. Completion steps and tokens are averaged over all task instances}
\label{tab:results}
\end{table*}

Table \ref{tab:results} shows the quantitative results of different types of agents performing each kitchen task.

The action-only baseline successfully completes all tasks but one, where it assumes \texttt{find a/an mug} is equivalent to \texttt{find a/an cup}, and ends the search pre-maturely without exploring the sink where the mug is actually placed.
On the other hand, our bootstrapped agent is able to finish most tasks completely using its learned production rule.
The only exceptions are when it is tasked to find an object that does not exist in the scene, which is not part of its training.
But with very limited queries, the bootstrapped agent is able to successfully complete those tasks as well.
This shows that the knowledge in the bootstrapped agent can be easily transferred to new objects in new environments.

The success rate and number of query tokens show two advantages of our framework.
First, it is verifiable such that it won't make false assumptions (e.g., confusing mugs with cups).
Second, it is much more efficient to be deployed into new environments as the production rules it learned can be easily transferred and require minimal further assistance from the LLM, saving computations and costs.

\begin{figure}[t]
    \centering
    \begin{subfigure}[b]{0.22\textwidth}
        \centering
        \includegraphics[width=\textwidth]{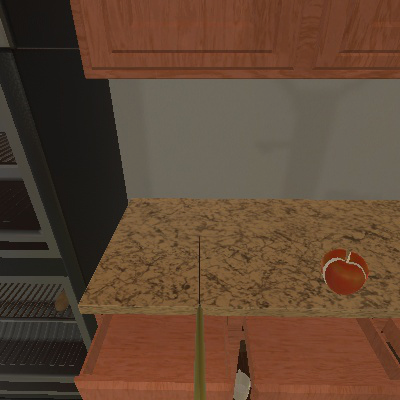}
        \caption{bootstrapped slicing}
        \label{fig:agent_slice}
    \end{subfigure}
    \hfill
    \begin{subfigure}[b]{0.22\textwidth}
        \centering
        \includegraphics[width=\textwidth]{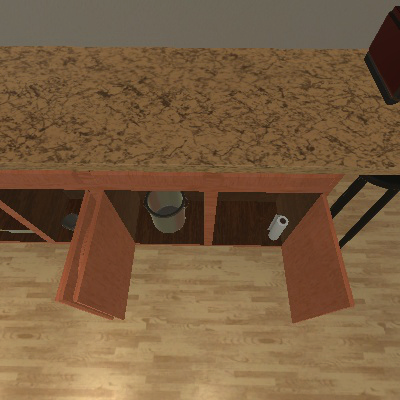}
        \caption{bootstrapped clearing}
        \label{fig:agent_clear}
    \end{subfigure}
    \\
    \begin{subfigure}[b]{0.22\textwidth}
        \centering
        \includegraphics[width=\textwidth]{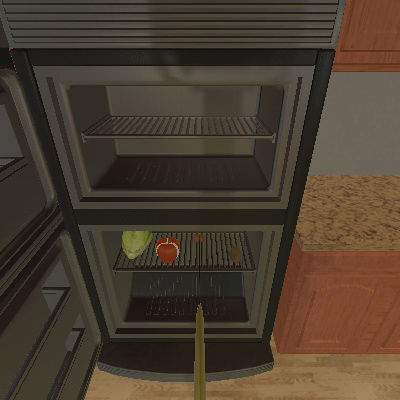}
        \caption{action-only slicing}
        \label{fig:action_slice}
    \end{subfigure}
    \hfill
    \begin{subfigure}[b]{0.22\textwidth}
        \centering
        \includegraphics[width=\textwidth]{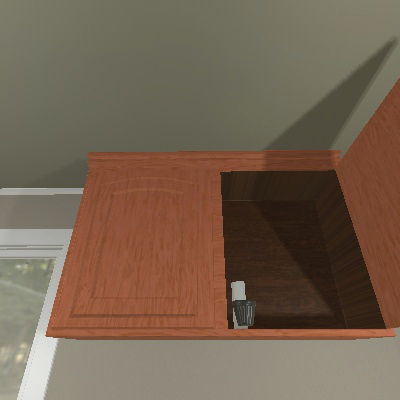}
        \caption{action-only clearing}
        \label{fig:action_clear}
    \end{subfigure}
    \caption{Examples of task execution. The first row shows the bootstrapped agent sliced the apple on the countertop, and put each object in their own cabinet. The second row shows the baseline agent sliced the apple at its current location (fridge), and put multiple objects in the same cabinet.}
    \label{fig:cases}
\end{figure}

We use paired sample t-test for means to compare the number of steps taken by both agents.
No significant evidence suggests that the two agents perform differently in \texttt{find} nor \texttt{slice} task (p-values $0.446$ and $0.347$).
This is not surprising as the knowledge source of both agents is the same LLM.

However, the bootstrapped agent is taking longer in the clearing task with significance (p-value $0.001$), which results from a stylistic difference between the two agents.
As shown in Figures \ref{fig:agent_clear} and \ref{fig:action_clear}, the bootstrapped agent places everything into an individual cabinet while the baseline places multiple objects in the same cabinets.
This is because one of the productions generated is ``if there is an object on the countertop and there is an empty receptacle, attend to the subtask pick up the object and place it into the empty receptacle''.
This production gets reused repeatedly, requiring the agent to seek an unique empty receptacle before placing each object instead of putting every object in the same cabinet.
By contrast, the baseline agent is making decisions on a case-by-case basis, so it does not enforce that the target receptacle has to be empty.

A similar difference is also found in the \texttt{slice} task where the bootstrapped agent always moves the objects to the countertops before slicing while the baseline agent slices objects at their current location (Figures \ref{fig:agent_slice} and \ref{fig:action_slice}).

\subsection{Production Analysis}

The following are some learned productions:
\begin{itemize}
    \item IF the current task is to find a/an \texttt{<object>} AND the \texttt{<object>} is located on \texttt{<location>} AND the robot is not at \texttt{<location>} THEN choose motor action: move to \texttt{<location>}.
    \item IF the current task is to slice a/an \texttt{<sliceable>} AND the robot is holding a/an \texttt{<sliceable>} AND there is no \texttt{<tool>} in the spatial knowledge or object knowledge THEN choose 'attend to subtask: find a/an \texttt{<tool>}'.
    \item IF the current task is to clear objects from a/an \texttt{<receptacle\_type>} AND all the \texttt{<receptacle\_type>} are empty THEN choose special action: 'done'.
\end{itemize}

These show that the agent is able to represent different aspects of the given tasks using production rules.
The first represents a common strategy for finding things, namely how to find things with a known location.
The second represents decomposing complex tasks and reusing previously learned tasks.
The third is a correct termination condition, which is not directly provided, for the exploration task from the LLM.

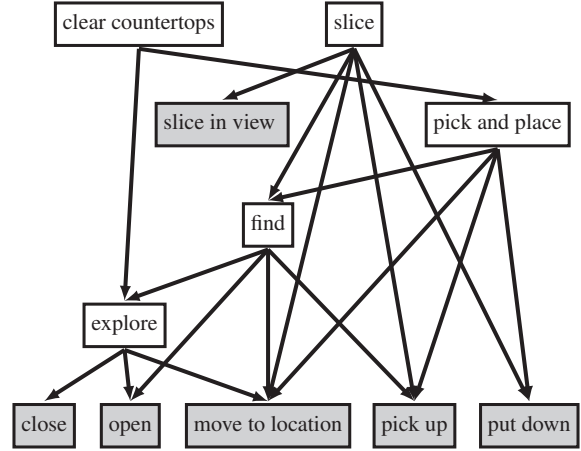
\begin{figure}[ht]
    \centering
    \resizebox{0.90\columnwidth}{!}{%
    \begin{tikzpicture}[
        graynode/.style={rectangle, draw=black, fill=gray!40!white, very thick, minimum size=0.08\columnwidth, scale=2.2, line width=3pt, },
        whitenode/.style={rectangle, draw=black, fill=white, very thick, minimum size=0.08\columnwidth, scale=2.2, line width=3pt, },
    ]
        
        \node[graynode] (close) at (1.2, 1) { close\strut };
        \node[graynode] (open) at (4.2, 1) { open\strut };
        \node[graynode] (move) at (9, 1) { move to location\strut };
        \node[graynode] (pick) at (14.1, 1) { pick up\strut };
        \node[graynode] (put) at (18.1, 1) { put down\strut };

        \node[whitenode] (explore) at (4, 4.5) { explore\strut };
        \node[whitenode] (find) at (9, 8) { find\strut };
        \node[graynode] (sliceb) at (7.4, 11.5) { slice in view \strut };
        \node[whitenode] (pandp) at (17, 11.5) { pick and place\strut };
        \node[whitenode] (clear) at (4.5, 15) { clear countertops\strut };
        \node[whitenode] (slice) at (12, 15) { slice\strut };

        \draw[-latex, line width=4pt] (explore.south) -- (close.north);
        \draw[-latex, line width=4pt] (explore.south) -- (open.north);
        \draw[-latex, line width=4pt] (explore.south) -- (move.north);
        \draw[-latex, line width=4pt] (find.south) -- (explore.north);
        \draw[-latex, line width=4pt] (find.south) -- (move.north);
        \draw[-latex, line width=4pt] (find.south) -- (pick.north);
        \draw[-latex, line width=4pt] (find.south) -- (open.north);
        \draw[-latex, line width=4pt] (pandp.south) -- (find.north);
        \draw[-latex, line width=4pt] (pandp.south) -- (pick.north);
        \draw[-latex, line width=4pt] (pandp.south) -- (put.north);
        \draw[-latex, line width=4pt] (pandp.south) -- (move.north);
        \draw[-latex, line width=4pt] (slice.south) -- (pick.north);
        \draw[-latex, line width=4pt] (slice.south) -- (find.north);
        \draw[-latex, line width=4pt] (slice.south) -- (move.north);
        \draw[-latex, line width=4pt] (slice.south) -- (sliceb.north);
        \draw[-latex, line width=4pt] (slice.south) -- (put.north);
        \draw[-latex, line width=4pt] (clear.south) -- (pandp.north);
        \draw[-latex, line width=4pt] (clear.south) -- (explore.north);
    \end{tikzpicture}
    }
    \caption{The hierarchy of tasks learned. Gray nodes denote the built-in functions of the robot, and white nodes represent the tasks learned from the curriculum. For built-in actions that involve an object (e.g., close), the object has to be within the field of view for the action to be taken. Special actions (i.e., done and quit) are omitted due to space constraints.}
    \label{fig:subtasks}
\end{figure}

Figure \ref{fig:subtasks} shows the task hierarchy learned by the agent after training on the given curriculum.
It shows how previously learned tasks are used to perform new tasks.
This reduces the number of queries needed for the LLM, fosters generality, and ensures the scalability of our approach.


\section{Discussion}

\subsection{Explainability}

Our framework touches upon all three aspects of explainability as defined by \citeauthor{milani2022survey} (\citeyear{milani2022survey}).
The preconditions of the productions directly specify the feature that is being used (feature importance).
Each production rule corresponds to a specific scenario during the bootstrapping process when it is created, which helps determine the training points that influence the learned policy (learning process).
Lastly, the production application process can be easily converted to a verifiable decision tree by merging the precondition checks of productions (policy-level explainability).

\subsection{Limitations}

In this work, we explore only the high-level decision-making process of the agent and rely heavily on having a well-defined interface for low-level actions, such as navigation and object manipulation.
There will likely be a considerable sim-to-real gap when applying this to physical agents.

Additionally, the English description generation step requires the decision-making process to be articulable to be converted to production rules.
This is hard for skills that cannot be fully expressed using language (e.g., sculpting).


\subsection{Future Work}


There are more learning opportunities in cognitive architectures such as updating the preconditions of productions or using separate productions for conflict resolutions.
These would help better extract the existing knowledge from LLMs to fit the specific agent and environment configurations.

Additionally, it is well-acknowledged that human values and preferences are hard to represent with a single reward function \cite{casper2023open}.
But the production rules are interpretable and can be easily modified to suit each individual without extensive computation.
Therefore it would interesting to examine whether this framework will facilitate personalization in human-AI collaboration tasks.


\section{Conclusion}

This paper presents a framework for bootstrapping a cognitive architecture from the existing noisy knowledge in LLMs, with minimal human inputs.
We demonstrated how such an agent could efficiently learn to perform kitchen tasks and be applied to new environments.
This work generalizes using LLMs to generate plans and provides an alternative to purely data-driven foundation models.
And finally, we shed light on how it will benefit personalized agents in the future.


\bibliography{aaai24}

\appendix
\onecolumn
\section{Technical Appendix}

\subsection{Step-by-Step Example of Learning One Production Rule}

This section shows an example of learning a new production rule for \texttt{slice a/an <object>} task.
In the interest of space, only relevant information is kept.
The original complete prompt, along with the responses from the LLM, are provided in the code and data supplementary material.

\subsubsection{System prompt for action selection and production description in English}

The system prompt mainly describes the robot's affordance model and explains the input of future user prompts.
It is the same prompt for all action selections.

\subsubsection{User prompt for action selection}
As mentioned in the methods section, the production generation is grounded to a specific instance, and the LLM is first asked to choose an action. 
The user prompt for action selection has a fixed template, where the information will be dynamically filled according to the actual knowledge of the agent.
Below is an example of the information provided in the prompt.
Unimportant information and static instructions are omitted in the interest of space.
The user prompt below and the system prompt will be provided to the LLM in the same request for the action selection.

\begin{verbatim}
[Current Task] slice a/an lettuce

[Current Location] in front of SinkBasin_28084e25

[Spatial Knowledge]
 * (0.0 meters away) RobotGripper(Gripper) has Lettuce_895e9ec5(Lettuce), and 
nothing  else
 * (0.0 meters away) SinkBasin_28084e25(SinkBasin) has been fully explored: it has 
Cup_26e78d79(Cup), Egg_113844f2(Egg), and nothing else
 * (0.9 meters away) CounterTop4(CounterTop) has been explored: it is empty
 * <<OTHER RECEPTACLES OMITTED>>
 
[Object Knowledge]
 * Apple_969aaa5d(Apple) at Fridge_683c888d
 * Cabinet_2fee59ce(Cabinet): opened,
 * Faucet_63e88a3b(Faucet):turned off,
 * Lettuce_895e9ec5(Lettuce) at RobotGripper
 * <<OTHER OBJECTS OMITTED>>
 
[Previous Tasks]
 * go to explore a/an sinkbasin_28084e25: True
 * find a/an lettuce: True
 * <<OTHER TASKS OMITTED>>
 
[Action History]
 * <<OTHER HISTORY OMITTED>>
 * (time 20) motor action: pick up lettuce (purpose: IF the current task is to find 
a/an <object> AND the <object> is located in a/an <receptacle> AND the robot is in 
front of the <receptacle> AND the robot's gripper is empty THEN choose motor action: 
pick up <object>.)
 * (time 21) special action: done (purpose: IF the current task is to find a/an 
<object> AND the robot's gripper has <object> THEN choose special action: 'done'.)

[Possible Options]
 * attend to subtask: find a/an <object> (Apply anytime. End condition: the robot 
has found the object and has it in its gripper.)
 * motor action: put Lettuce_895e9ec5 on CounterTop4
 * special action: 'done'
 * <<OTHER OPTIONS OMITTED>>
 
[Blacklisted Options]
 * attend to subtask: slice a/an lettuce
 
\end{verbatim}

\subsubsection{GPT4 response for action selection} The response of GPT4 to the prompt above is recreated below.

\begin{verbatim}
[Current Task]
Slice a lettuce.

[Common Strategies for "slice a lettuce"]
 * Find a lettuce and a knife.
 * Pick up the lettuce.
 * Move to a countertop or a similar flat surface.
 * Put down the lettuce.
 * Pick up the knife.
 * Slice the lettuce.

[Goal Conditions]
 * The lettuce is sliced.

[Current Situation Analysis]
The robot is currently holding a lettuce and is located in front of the sink. The 
robot has explored several areas of the kitchen, including the sink, countertop4, 
countertop5, and a few cabinets. However, there are still several unexplored areas,
including countertop1, countertop2, countertop3, and several cabinets. The robot has
not found a knife yet, which is necessary for slicing the lettuce. The robot's 
previous actions indicate that it has successfully found and picked up the lettuce,
which is part of the task. There is no existing plan provided.

[Task Analysis]
The current task is to slice a lettuce. The robot has already found and picked up
the lettuce, which is part of the task. However, the robot has not found a knife 
yet, which is necessary for slicing the lettuce. The robot also needs to find a 
suitable place to put down the lettuce for slicing, such as a countertop.

[Plan]
Step1: Put down the lettuce on a countertop.
Step2: Find and pick up a knife.
Step3: Slice the lettuce.

[Option Evaluation]
"attend to subtask: find a/an <object>": This option is not useful right now because 
the robot already has the lettuce in its gripper.
"motor action: put Lettuce_895e9ec5 on CounterTop4": This option is useful because 
it allows the robot to free its gripper so it can pick up a knife.
"special action: 'done'": This option is not appropriate because the task is not 
completed yet.
<<OTHER EVALUATION OMITTED>>

[Option Suggestion]
"motor action: put Lettuce_895e9ec5 on CounterTop4"

[Purpose]
The purpose of the suggested option is to free the robot's gripper so it can pick up 
a knife.

[End]
\end{verbatim}

\subsubsection{User prompt for generating English description of the production rule}
After parsing the suggested option from the previous response, we prompt the LLM again for the English description of the production rule.
We combine all the previous queries and responses and send all of them to the LLM in a single request.

\subsubsection{GPT4 response for production rule in English}

\begin{verbatim}
[Relevant Information]
 * The current task is to slice a lettuce.
 * The robot is currently holding a lettuce in its gripper.
 * The robot is currently in front of the sink, which is not an ideal place for 
slicing.
 * A countertop is a suitable place for slicing.
 * The robot has the option to put the lettuce on CounterTop4.

[Specific Rule]
IF the current task is to slice a lettuce AND the robot is holding a lettuce in its 
gripper AND the robot is not at a suitable place for slicing AND there is a 
countertop available THEN choose motor action: put Lettuce_895e9ec5 on CounterTop4.

[Generalizable Constants]
 * Lettuce_895e9ec5 could be any object that needs to be sliced.
 * CounterTop4 could be any countertop or similar flat surface.

[Generalized Rule]
IF the current task is to slice a/an <object> AND the robot is holding the <object> 
in its gripper AND the robot is not at a suitable place for slicing AND there is a 
countertop available THEN choose motor action: put <object> on <countertop>.

[Correspondence]
 * <object> is Lettuce_895e9ec5.
 * <countertop> is CounterTop4.

Notes: This rule is designed to ensure that the robot places the object to be sliced 
on a suitable surface before attempting to slice it. It is applicable to any task 
that involves slicing an object.
\end{verbatim}

\subsubsection{System prompt for production rule code generation}

The system prompt mainly describes the Python interfaces for the productions and information stored in the agent.
It is the same prompt for all code generation.

\subsubsection{User prompt for production rule code generation}

The response of the production rule in English is copied verbatim along with the instructions.

\subsubsection{GPT4 response for production code generation} The response generated by GPT4 is recreated below. The code has been reformatted to fit the page.

\begin{verbatim}
[Variable Bindings]
 * target_object: the object that needs to be sliced. Its value should be parsed 
from the current task.
 * target_receptacle: a suitable place for slicing. Its value should be assigned by 
iterating through all receptacles and finding the countertop that is closest to the 
robot.

[Precondition Relation]
 * "the current task is to slice a/an <object>" is about the current task
 * "the robot is holding the <object> in its gripper" is about the object type of 
target_object and the hosting attribute of the robot's gripper
 * "the robot is not at a suitable place for slicing" is about the current location
 * "there is a countertop available" is about the existence of target_receptacle

[Precondition Test]
 * The current task is to slice a/an <object> can be tested by checking if 
`re.match(r"slice a/an (.+)", current_task)`
 * The robot is holding the <object> in its gripper can be tested by checking if 
`spatial_knowledge["RobotGripper"].hosting and name_equal(
spatial_knowledge["RobotGripper"].hosting[0].object_type, target_object)`
 * The robot is not at a suitable place for slicing can be tested by checking if 
`not OracleQuery(f"{current_location} is a suitable place for slicing")`
 * There is a countertop available can be tested by checking if `target_receptacle
is not None`

[Python Implementation]
```py
import re
from typing import Dict, Tuple
from .knowledge import SpatialKnowledge, ObjectKnowledge

class SliceObject:

    target_task: str = "slice a/an <sliceable>"

    def precondition(
        self, 
        current_task: str,
        current_location: str,
        previous_tasks: Dict[str, bool],
        spatial_knowledge: SpatialKnowledge, 
        object_knowledge: ObjectKnowledge,
    ) -> Tuple[bool, str]:
        match = re.match(r"slice a/an (.+)", current_task)
        if not match:
            return False, "The current task is not to slice an object"
        target_object = match.group(1)

        if not spatial_knowledge["RobotGripper"].hosting or not name_equal(
            spatial_knowledge["RobotGripper"].hosting[0].object_type, 
            target_object,
        ):
            return (
                False, 
                "The robot is not holding the target object in its gripper",
            )

        if OracleQuery(f"{current_location} is a suitable place for slicing"):
            return False, "The robot is already at a suitable place for slicing"

        target_receptacle = min(
            (
                r for r in spatial_knowledge.locations.values() 
                if name_equal(r.object_type, "CounterTop")
            ), 
            key=spatial_knowledge.get_distance
        )
        if target_receptacle is None:
            return False, "There is no countertop available"

        setattr(self, "target_object", target_object)
        setattr(self, "target_receptacle", target_receptacle.object_name)
        return True, ""

    def apply(self) -> str:
        return f"motor action: put {self.target_object} on {self.target_receptacle}"
``` 
\end{verbatim}

\subsubsection{User prompt for code revision}

As described in the production improvement section, if the code generated does not pass the current state knowledge, we query it again with the current knowledge and the error raised, which in this example is the world knowledge cannot determine whether ``SinkBasin\_28084e25 is a suitable place for slicing'' is true or not.

\begin{verbatim}
<<INFORMATION FROM THE ACTION SELECTION STEP>>

[Expected Return of `precondition()`]
(True, "")

[Your Code Return]
ValueError: got unknown statement: SinkBasin_28084e25 is a suitable place for 
slicing
\end{verbatim}

\subsubsection{GPT4 response for code revision}

\begin{verbatim}
[Error Interpretation]
The message indicates that the OracleQuery function has failed.

[Error Attribution]
The error is because the OracleQuery function does not understand the statement 
"SinkBasin_28084e25 is a suitable place for slicing". The OracleQuery function is 
designed to answer general knowledge questions, not specific queries about the 
current environment.

[Things to Change]
The OracleQuery test should be replaced with a check against a list of suitable 
places for slicing.

[Python Implementation]
```py
import re
from typing import Dict, Tuple
from .knowledge import SpatialKnowledge, ObjectKnowledge

class SliceObject:

    target_task: str = "slice a/an <sliceable>"

    def precondition(
        self, 
        current_task: str,
        current_location: str,
        previous_tasks: Dict[str, bool],
        spatial_knowledge: SpatialKnowledge, 
        object_knowledge: ObjectKnowledge,
    ) -> Tuple[bool, str]:
        match = re.match(r"slice a/an (.+)", current_task)
        if not match:
            return False, "The current task is not to slice an object"
        target_object = match.group(1)

        if not spatial_knowledge["RobotGripper"].hosting or not name_equal(
            spatial_knowledge["RobotGripper"].hosting[0].object_type, 
            target_object,
        ):
            return (
                False, 
                "The robot is not holding the target object in its gripper",
            )

        suitable_places_for_slicing = ["CounterTop"]
        if not any(
            name_equal(current_location, place) 
            for place in suitable_places_for_slicing
        ):
            return False, "The robot is not at a suitable place for slicing"

        target_receptacle = min(
            (
                r for r in spatial_knowledge.locations.values() 
                if name_equal(r.object_type, "CounterTop")
            ), 
            key=spatial_knowledge.get_distance,
        )
        if target_receptacle is None:
            return False, "There is no countertop available"

        setattr(self, "target_object", target_object)
        setattr(self, "target_receptacle", target_receptacle.object_name)
        return True, ""

    def apply(self) -> str:
        return f"motor action: put {self.target_object} on {self.target_receptacle}"
```
\end{verbatim}

\subsection{Production Rules Generated}

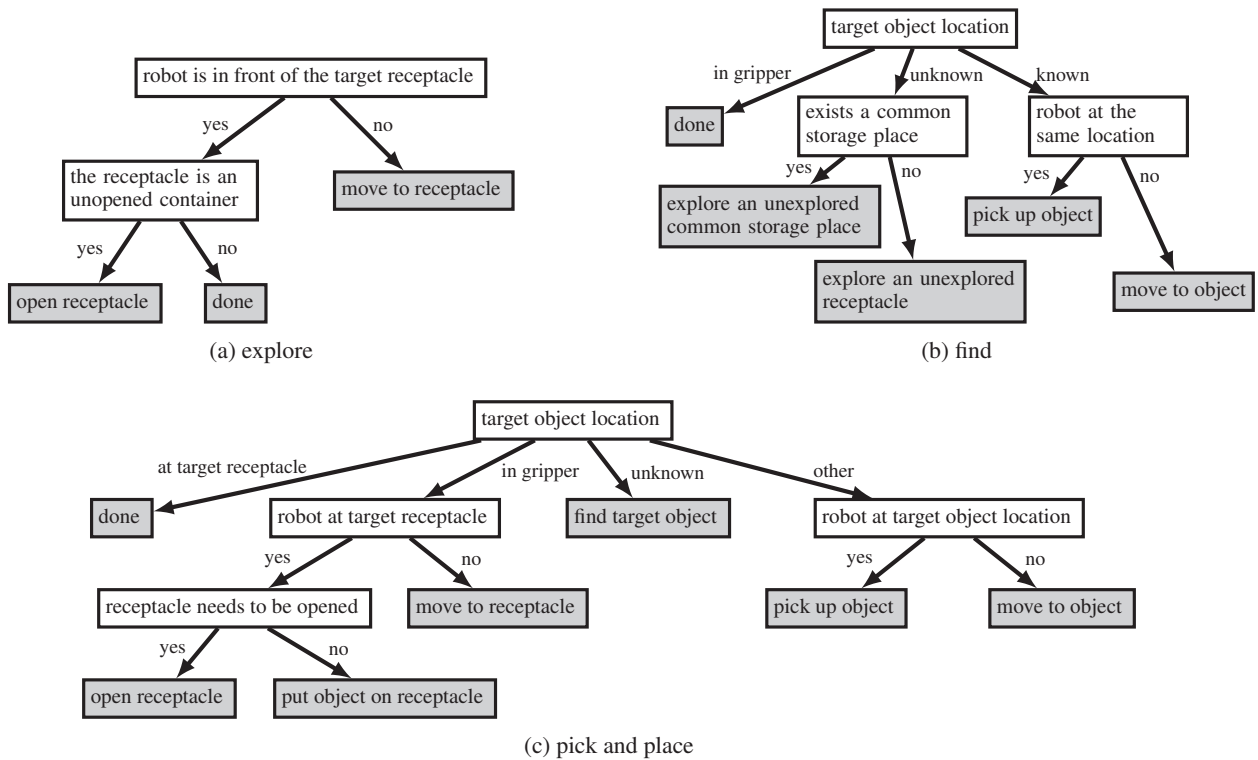
\begin{figure}[h]
    \centering
    \begin{subfigure}[b]{0.48\textwidth}
        \centering
        \begin{tikzpicture}[
            graynode/.style={rectangle, draw=black, fill=gray!40!white, very thick, scale=0.8},
            whitenode/.style={rectangle, draw=black, fill=white, very thick, scale=0.8},
        ]
        \node[whitenode] (explore_feature_location) at (4.5, -0.5) { robot is in front of the target receptacle \strut };
        \node[graynode] (explore_action_move) at (6, -2) { move to receptacle \strut };
        \node[whitenode, text width = 0.35\textwidth] (explore_feature_explored) at (2.5, -2) { the receptacle is an unopened container \strut };
        \node[graynode] (explore_action_open) at (1.5, -3.5) { open receptacle \strut };
        \node[graynode] (explore_action_done) at (3.5, -3.5) { done \strut };
        
        \draw[-latex, ultra thick] (explore_feature_location) -- (explore_feature_explored) node [pos=0.45, left, scale=0.75] { yes$~~$\strut };
        \draw[-latex, ultra thick] (explore_feature_location) -- (explore_action_move) node [pos=0.4, right, scale=0.75] { $~$no\strut };
        \draw[-latex, ultra thick] (explore_feature_explored) -- (explore_action_open) node [pos=0.45, left, scale=0.75] { yes$~~$\strut };
        \draw[-latex, ultra thick] (explore_feature_explored) -- (explore_action_done) node [pos=0.45, right, scale=0.75] { $~~$no\strut };
        \end{tikzpicture}
        \caption{explore}
        \label{fig:productions_explore}
    \end{subfigure}
    \hfill
    \begin{subfigure}[b]{0.48\textwidth}
        \centering
        \begin{tikzpicture}[
            graynode/.style={rectangle, draw=black, fill=gray!40!white, very thick, scale=0.8},
            whitenode/.style={rectangle, draw=black, fill=white, very thick, scale=0.8},
        ]
        \node[whitenode] (find_feature_location) at (3.5, -0.5) { target object location\strut };
        \node[graynode] (find_action_done) at (0.5, -1.8) { done\strut };
        \node[whitenode, text width = 0.3\textwidth] (find_feature_common) at (3, -1.8) { exists a common storage place\strut };
        \node[whitenode, text width = 0.28\textwidth] (find_feature_robot) at (6, -1.8) { robot at the same location\strut };
        \node[graynode, text width = 0.40\textwidth] (find_action_explore_common) at (1.5, -3) { explore an unexplored\strut \\ common storage place\strut };
        \node[graynode, text width = 0.38\textwidth] (find_action_explore_recep) at (3.5, -4) { explore an unexplored\strut \\ receptacle\strut };
        \node[graynode] (find_action_pick) at (5, -3) { pick up object\strut };
        \node[graynode] (find_action_move) at (7, -4) { move to object\strut };
    
        \draw[-latex, ultra thick] (find_feature_location) -- (find_action_done) node [pos=0.4, left, scale=0.75] { in gripper$~~~$\strut };
        \draw[-latex, ultra thick] (find_feature_location) -- (find_feature_common) node [pos=0.6, right, scale=0.75] { unknown\strut };
        \draw[-latex, ultra thick] (find_feature_location) -- (find_feature_robot) node [pos=0.6, right, scale=0.75] { $~~~$known\strut };
        \draw[-latex, ultra thick] (find_feature_common) -- (find_action_explore_common) node [pos=0.4, left, scale=0.75] { yes$~~$\strut };
        \draw[-latex, ultra thick] (find_feature_common) -- (find_action_explore_recep) node [pos=0.15, right, scale=0.75] { no\strut };
        \draw[-latex, ultra thick] (find_feature_robot) -- (find_action_pick) node [pos=0.45, left, scale=0.75] { yes$~~$\strut };
        \draw[-latex, ultra thick] (find_feature_robot) -- (find_action_move) node [pos=0.18, right, scale=0.75] { no\strut };
        \end{tikzpicture}
        \caption{find}
        \label{fig:productions_find}
    \end{subfigure} \\
    \bigskip
    \begin{subfigure}[b]{0.98\textwidth}
        \centering
        \begin{tikzpicture}[
            graynode/.style={rectangle, draw=black, fill=gray!40!white, very thick, scale=0.8},
            whitenode/.style={rectangle, draw=black, fill=white, very thick, scale=0.8},
        ]
        \node[whitenode] (pandp_feature_location) at (7, -0.5) { target object location \strut };
        \node[graynode] (pandp_action_done) at (1, -1.8) { done \strut };
        \node[whitenode] (pandp_feature_robot_recep) at (4.5, -1.8) { robot at target receptacle \strut };
        \node[graynode] (pandp_action_find) at (8, -1.8) { find target object \strut };
        \node[whitenode] (pandp_feature_robot_object) at (12, -1.8) { robot at target object location \strut };
        \node[whitenode] (pandp_feature_open) at (2.5, -3) { receptacle needs to be opened \strut };
        \node[graynode] (pandp_action_move) at (6, -3) { move to receptacle \strut };
        \node[graynode] (pandp_action_pick) at (10.5, -3) { pick up object \strut };
        \node[graynode] (pandp_action_move2) at (13.5, -3) { move to object \strut };
        \node[graynode] (pandp_action_open) at (1.5, -4.2) { open receptacle \strut };
        \node[graynode] (pandp_action_put) at (4.5, -4.2) { put object on receptacle \strut };
        
        \draw[-latex, ultra thick] (pandp_feature_location) -- (pandp_action_done) node [pos=0.4, left, scale=0.75] { at target receptacle $~~~~~~$\strut };
        \draw[-latex, ultra thick] (pandp_feature_location) -- (pandp_feature_robot_recep) node [pos=0.55, right, scale=0.75] { $~~~~$in gripper };
        \draw[-latex, ultra thick] (pandp_feature_location) -- (pandp_action_find) node [pos=0.55, right, scale=0.75] { $~~$unknown };
        \draw[-latex, ultra thick] (pandp_feature_location) -- (pandp_feature_robot_object) node [pos=0.52, right, scale=0.75] { $~~~~~~~~$other };
        \draw[-latex, ultra thick] (pandp_feature_robot_recep) -- (pandp_feature_open) node [pos=0.45, left, scale=0.75] { yes$~~~$ };
        \draw[-latex, ultra thick] (pandp_feature_robot_recep) -- (pandp_action_move) node [pos=0.45, right, scale=0.75] { $~~~$no };
        \draw[-latex, ultra thick] (pandp_feature_robot_object) -- (pandp_action_pick) node [pos=0.45, left, scale=0.75] { yes$~~~$ };
        \draw[-latex, ultra thick] (pandp_feature_robot_object) -- (pandp_action_move2) node [pos=0.45, right, scale=0.75] { $~~~$no };
        \draw[-latex, ultra thick] (pandp_feature_open) -- (pandp_action_open) node [pos=0.45, left, scale=0.75] { yes$~$ };
        \draw[-latex, ultra thick] (pandp_feature_open) -- (pandp_action_put) node [pos=0.45, right, scale=0.75] { $~~~$no };
        \end{tikzpicture}
        \caption{pick and place}
        \label{fig:productions_place}
    \end{subfigure}
    \caption{Productions learned through bootstrapping for exploring, finding, and placing. Gray nodes are the effects, and white nodes are the features being conditioned on.}
    \label{fig:productions1}
\end{figure}

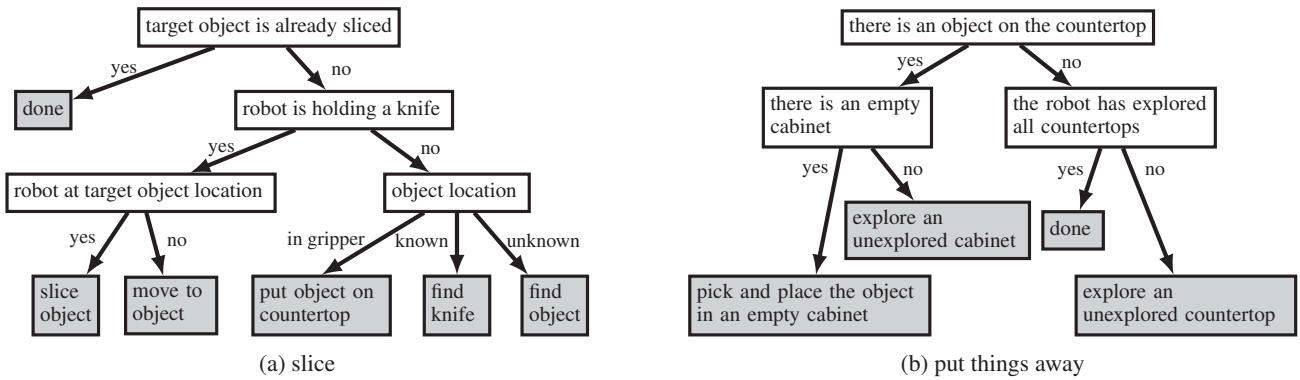
\begin{figure}[ht]
    \begin{subfigure}[b]{0.48\textwidth}
        \centering
        \begin{tikzpicture}[
            graynode/.style={rectangle, draw=black, fill=gray!40!white, very thick, scale=0.8},
            whitenode/.style={rectangle, draw=black, fill=white, very thick, scale=0.8},
        ]
        
        \node[whitenode] (slice_feature_sliced) at (3.5, -0.5) { target object is already sliced\strut };
        \node[graynode] (slice_action_done) at (0.5, -1.6) { done\strut };
        \node[whitenode] (slice_feature_hold) at (4.5, -1.6) { robot is holding a knife \strut };
        \node[whitenode] (slice_feature_location) at (1.8, -2.7) { robot at target object location \strut };
        \node[whitenode] (slice_feature_object) at (6, -2.7) { object location \strut };
        \node[graynode, text width = 0.10\textwidth] (slice_action_slice) at (0.8, -4.2) { slice object\strut };
        \node[graynode, text width = 0.15\textwidth] (slice_action_move) at (2.2, -4.2) { move to object\strut };
        \node[graynode, text width = 0.24\textwidth] (slice_action_put) at (4.2, -4.2) { put object on countertop\strut };
        \node[graynode, text width = 0.10\textwidth] (slice_action_knife) at (6, -4.2) { find knife\strut };
        \node[graynode, text width = 0.10\textwidth] (slice_action_object) at (7.3, -4.2) { find object\strut };

        \draw[-latex, ultra thick] (slice_feature_sliced) -- (slice_action_done) node [pos=0.45, left, scale=0.75] { yes$~$ };
        \draw[-latex, ultra thick] (slice_feature_sliced) -- (slice_feature_hold) node [pos=0.55, right, scale=0.75] { $~~~$no };
        \draw[-latex, ultra thick] (slice_feature_hold) -- (slice_feature_location) node [pos=0.45, left, scale=0.75] { yes$~$ };
        \draw[-latex, ultra thick] (slice_feature_hold) -- (slice_feature_object) node [pos=0.45, right, scale=0.75] { $~~~$no };
        \draw[-latex, ultra thick] (slice_feature_location) -- (slice_action_slice) node [pos=0.45, left, scale=0.75] { yes$~$ };
        \draw[-latex, ultra thick] (slice_feature_location) -- (slice_action_move) node [pos=0.45, right, scale=0.75] { $~$no };
        \draw[-latex, ultra thick] (slice_feature_object) -- (slice_action_put.north) node [pos=0.45, left, scale=0.75] { in gripper$~$ };
        \draw[-latex, ultra thick] (slice_feature_object) -- (slice_action_knife) node [pos=0.45, left, scale=0.75] { known };
        \draw[-latex, ultra thick] (slice_feature_object) -- (slice_action_object) node [pos=0.45, right, scale=0.75] { unknown };
        \end{tikzpicture}
        \caption{slice}
        \label{fig:productions_slice}
    \end{subfigure}
    \hfill
    \begin{subfigure}[b]{0.48\textwidth}
        \centering
        \begin{tikzpicture}[
            graynode/.style={rectangle, draw=black, fill=gray!40!white, very thick, scale=0.8},
            whitenode/.style={rectangle, draw=black, fill=white, very thick, scale=0.8},
        ]
        \node[whitenode] (clear_feature_object) at (4, -0.5) { there is an object on the countertop };
        \node[whitenode, text width = 0.3\textwidth] (clear_feature_cabinet) at (2, -1.7) { there is an empty cabinet \strut };
        \node[whitenode, text width = 0.38\textwidth] (clear_feature_countertop) at (5.5, -1.7) { the robot has explored all countertops \strut };
        \node[graynode, text width = 0.44\textwidth] (clear_action_pick) at (1.5, -4.2) { pick and place the object in an empty cabinet \strut };
        \node[graynode, text width = 0.33\textwidth] (clear_action_cabinet) at (3.2, -3.2) { explore an \\ unexplored cabinet };
        \node[graynode] (clear_action_done) at (5, -3.2) { done \strut };
        \node[graynode, text width = 0.4\textwidth] (clear_action_countertop) at (6.5, -4.2) { explore an \\ unexplored countertop \strut };
        
        \draw[-latex, ultra thick] (clear_feature_object) -- (clear_feature_cabinet) node [pos=0.45, left, scale=0.75] { yes$~$ };
        \draw[-latex, ultra thick] (clear_feature_object) -- (clear_feature_countertop) node [pos=0.45, right, scale=0.75] { $~~$no };
        \draw[-latex, ultra thick] (clear_feature_cabinet) -- (clear_action_pick) node [pos=0.18, left, scale=0.75] { yes };
        \draw[-latex, ultra thick] (clear_feature_cabinet) -- (clear_action_cabinet) node [pos=0.43, right, scale=0.75] { $~$no };
        \draw[-latex, ultra thick] (clear_feature_countertop) -- (clear_action_done) node [pos=0.4, left, scale=0.75] { yes };
        \draw[-latex, ultra thick] (clear_feature_countertop) -- (clear_action_countertop) node [pos=0.18, right, scale=0.75] { $~$no };
        \end{tikzpicture}
        \caption{put things away}
        \label{fig:productions_clear}
    \end{subfigure}
    \caption{Productions learned through bootstrapping for slicing and putting things away. Gray nodes are the effects, and white nodes are the features being conditioned on.}
    \label{fig:productions2}
\end{figure}

Figures \ref{fig:productions1} and \ref{fig:productions2} illustrate the production rules learned for each task in the form of decision trees.
These decision trees are for illustrations purpose and are not completely equivalent to the agent because of the sampling mechanism in the production selection process.
The order of the nodes is chosen based on the utility of the production.
That is, productions with higher utility (e.g., productions leading to the \texttt{done} action) will be closer to the root.
This demonstrates how the production rules can be converted to verifiable decision trees as mentioned in the discussion section in the main paper.

The actual production rules and their Python implementations can be found in the code and data supplementary material.

\subsection{Tokens Usage}

\begin{figure}
    \centering
    \includegraphics[width=0.99\textwidth]{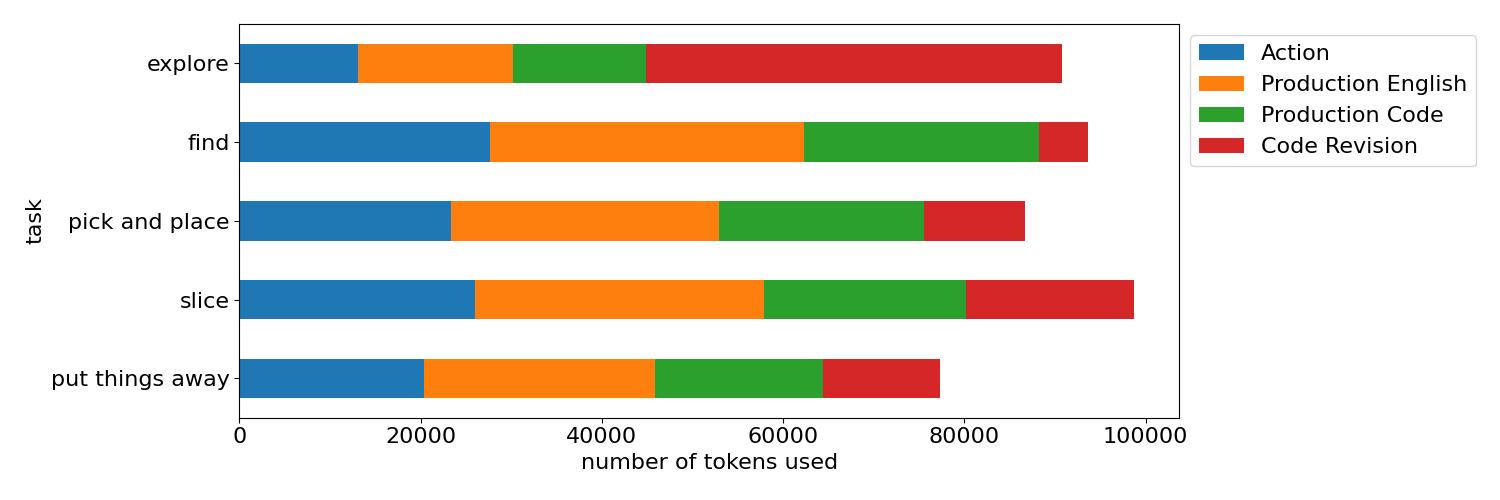}
    \caption{Number of tokens used during bootstrapping}
    \label{fig:tokens}
\end{figure}

As shown in Figure \ref{fig:tokens}, the number of tokens needed to train each task is roughly the same.
So as the curriculum expands, the number of tokens needed will only grow linearly.
Additionally, the number of tokens needed to train one task is less than one single trial of slicing objects of the action-only agent as reflected in Table \ref{tab:results}.
This shows that our framework is much more cost-effective.
The testing experiments on the baseline action-only agent cost around $\$120$ in total while the bootstrapping of our framework costs less than $\$40$.

\subsection{Step-by-Step Example of Completing One Task}

\begin{figure}[ht]
    \centering
    \begin{subfigure}[b]{0.16\textwidth}
        \centering
        \includegraphics[width=\textwidth]{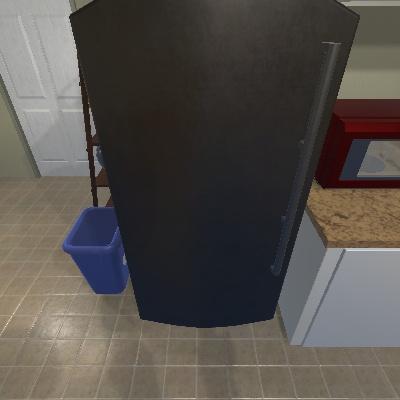}
        \caption{}
        \label{fig:sample_a}
    \end{subfigure}
    \hfill
    \begin{subfigure}[b]{0.16\textwidth}
        \centering
        \includegraphics[width=\textwidth]{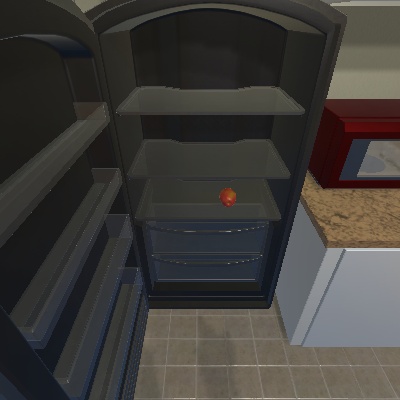}
        \caption{}
        \label{fig:sample_b}
    \end{subfigure}
    \hfill
    \begin{subfigure}[b]{0.16\textwidth}
        \centering
        \includegraphics[width=\textwidth]{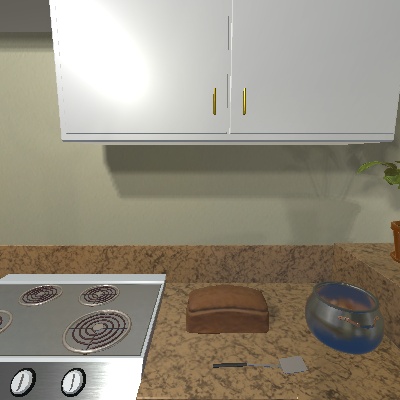}
        \caption{}
        \label{fig:sample_c}
    \end{subfigure}
    \hfill
    \begin{subfigure}[b]{0.16\textwidth}
        \centering
        \includegraphics[width=\textwidth]{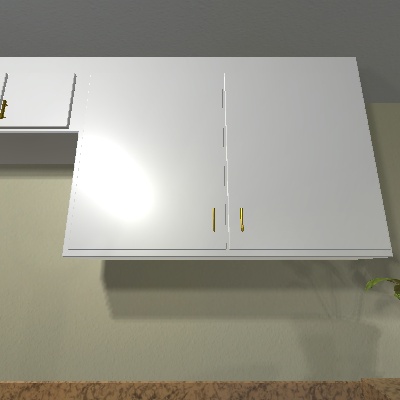}
        \caption{}
        \label{fig:sample_d}
    \end{subfigure}
    \hfill
    \begin{subfigure}[b]{0.16\textwidth}
        \centering
        \includegraphics[width=\textwidth]{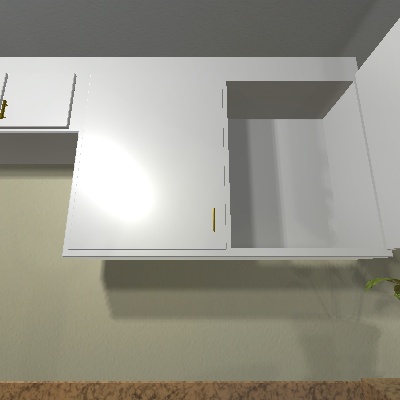}
        \caption{}
        \label{fig:sample_e}
    \end{subfigure}
    \hfill
    \begin{subfigure}[b]{0.16\textwidth}
        \centering
        \includegraphics[width=\textwidth]{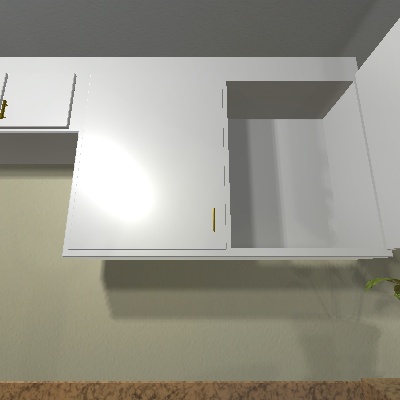}
        \caption{}
        \label{fig:sample_f}
    \end{subfigure}
    \\
    \smallskip
    \begin{subfigure}[b]{0.16\textwidth}
        \centering
        \includegraphics[width=\textwidth]{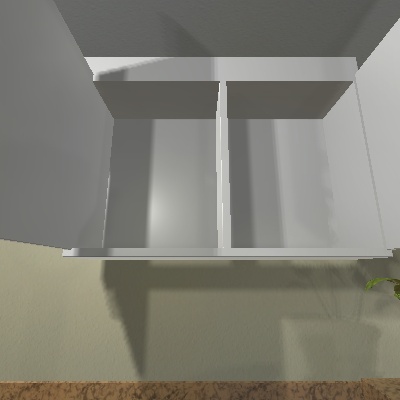}
        \caption{}
        \label{fig:sample_g}
    \end{subfigure}
    \hfill
    \begin{subfigure}[b]{0.16\textwidth}
        \centering
        \includegraphics[width=\textwidth]{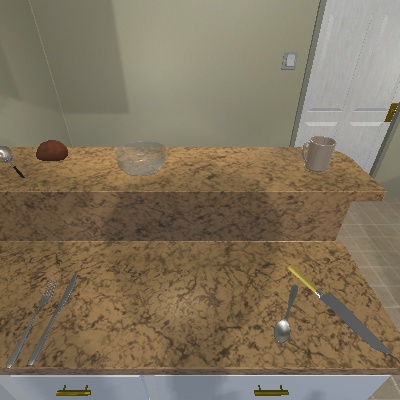}
        \caption{}
        \label{fig:sample_h}
    \end{subfigure}
    \hfill
    \begin{subfigure}[b]{0.16\textwidth}
        \centering
        \includegraphics[width=\textwidth]{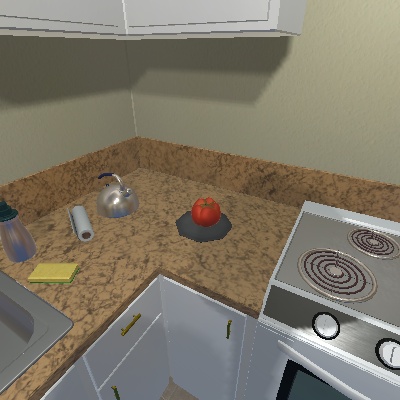}
        \caption{}
        \label{fig:sample_i}
    \end{subfigure}
    \hfill
    \begin{subfigure}[b]{0.16\textwidth}
        \centering
        \includegraphics[width=\textwidth]{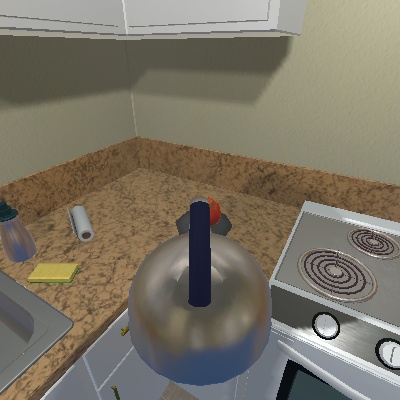}
        \caption{}
        \label{fig:sample_j}
    \end{subfigure}
    \hfill
    \begin{subfigure}[b]{0.16\textwidth}
        \centering
        \includegraphics[width=\textwidth]{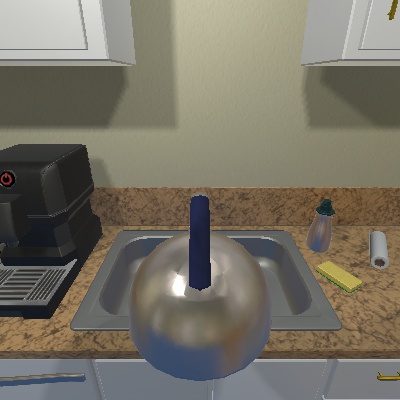}
        \caption{}
        \label{fig:sample_k}
    \end{subfigure}
    \hfill
    \begin{subfigure}[b]{0.16\textwidth}
        \centering
        \includegraphics[width=\textwidth]{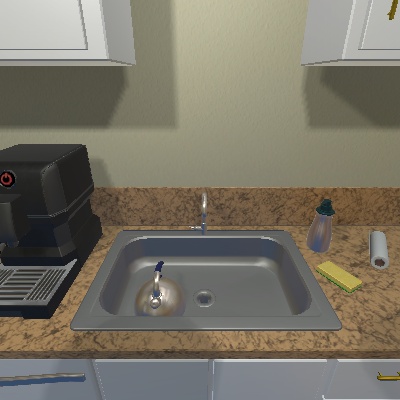}
        \caption{}
        \label{fig:sample_l}
    \end{subfigure}
    \caption{Trajectory of the agent completing ``pick up and place a/an kettle in/on a/an sinkbasin''}
    \label{fig:sample}
\end{figure}

\begin{table*}[ht]
    \centering
    \begin{tabular}{m{0.2\textwidth}|llm{0.48\textwidth}}
        Main Task & Figure & Action & Production Preconditions \\ \hline \hline
        pick and place a/an kettle in/on a/an sinkbasin & \ref{fig:sample_a} & find a/an kettle & the robot's gripper is empty AND the \texttt{<object>} has not been located \\ \hline
        \multirow{ 3}{*}{find a/an kettle} & \ref{fig:sample_a} & open fridge & the robot is in front of a closed \texttt{<receptacle>} AND the robot's gripper is empty \\ \cline{2-4}
        & \ref{fig:sample_b} & move to countertop2 & the robot's gripper is empty AND there is an unexplored \texttt{<receptacle>} that is commonly associated with the \texttt{<object>} \\ \cline{2-4}
        & \ref{fig:sample_c} & explore a/an cabinet1 & the robot's gripper is empty AND there are unexplored receptacles in the kitchen \\ \hline
        \multirow{ 3}{*}{explore a/an cabinet1} & \ref{fig:sample_c} & move to cabinet1 & the \texttt{<receptacle>} is not at the current location AND the robot's gripper is empty \\ \cline{2-4}
        & \ref{fig:sample_d} & open cabinet1 & the robot is at the location of the \texttt{<receptacle>} AND the \texttt{<receptacle>} is closed and unexplored AND the robot's gripper is empty \\ \cline{2-4}
        & \ref{fig:sample_e}& done & the robot is in front of the \texttt{<receptacle>} AND the \texttt{<receptacle>} has been fully explored \\ \hline
        find a/an kettle & \ref{fig:sample_e} & explore a/an cabinet2 & the robot's gripper is empty AND there are unexplored receptacles in the kitchen \\ \hline
        \multirow{ 3}{*}{explore a/an cabinet2} & \ref{fig:sample_e} & move to cabinet2 & the \texttt{<receptacle>} is not at the current location AND the robot's gripper is empty \\ \cline{2-4}
        & \ref{fig:sample_f} & open cabinet2 & the robot is at the location of the \texttt{<receptacle>} AND the \texttt{<receptacle>} is closed and unexplored AND the robot's gripper is empty \\ \cline{2-4}
        & \ref{fig:sample_g} & done & the robot is in front of the \texttt{<receptacle>} AND the \texttt{<receptacle>} has been fully explored \\ \hline
        \multirow{ 4}{*}{find a/an kettle} & \ref{fig:sample_g} & move to countertop1 & the robot's gripper is empty AND there is an unexplored \texttt{<receptacle>} that is commonly associated with the \texttt{<object>} \\ \cline{2-4}
        & \ref{fig:sample_h} & move to countertop3 & the robot's gripper is empty AND there is an unexplored \texttt{<receptacle>} that is commonly associated with the \texttt{<object>} \\ \cline{2-4}
        & \ref{fig:sample_i} & pick up kettle & the \texttt{<object>} is located in a/an \texttt{<receptacle>} AND the robot is in front of the \texttt{<receptacle>} \\ \cline{2-4}
        & \ref{fig:sample_j} & done & the robot's gripper has \texttt{<object>} \\ \hline
        \multirow{ 3}{0.2\textwidth}{pick and place a/an kettle in/on a/an sinkbasin} & \ref{fig:sample_j} & move to sinkbasin & the robot's gripper has the \texttt{<object>} AND the robot is not at the \texttt{<receptacle>} \\ \cline{2-4}
        & \ref{fig:sample_k} & put on sinkbasin & the robot is holding the \texttt{<object>} AND the robot is in front of the \texttt{<receptacle>} AND the \texttt{<receptacle>} is empty \\ \cline{2-4}
        & \ref{fig:sample_l} & done & the \texttt{<object>} is already in the \texttt{<receptacle>} AND the robot's gripper is empty \\ \hline
    \end{tabular}
    \caption{Action history of the agent completing ``pick up and place a/an kettle in/on a/an sinkbasin''. Due to space constraints, the object names are simplified and the task matching is omitted from the preconditions.}
    \label{tab:sample}
\end{table*}

Figure \ref{fig:sample} and Table \ref{tab:sample} shows the trajectory of the agent completing the task of ``pick up and place a/an kettle in/on a/an sinkbasin'' after bootstrapping.
The agent first attends to the subtask of finding a kettle, during which process it also uses the explore subtasks, and finally moves to the sink basin and places the kettle as instructed.
The main task column in the table reflects the management of the task stack in the agent: it attends to a single main task at a time and releases it when a production rule determines the current task is done.

\subsection{Task End Conditions Generated}

Here is a list of end conditions for the tasks families in our curriculum.
\begin{itemize}
    \item \texttt{explore a/an <receptacle>}: ``the robot has fully explored the receptacle.''
    \item \texttt{find a/an <object>}: ``the robot has found the object and has it in its gripper.''
    \item \texttt{pick up and place a/an <object> in/on a/an <receptacle>}: ``the robot has successfully picked up the specified object and placed it in/on the specified receptacle, and the robot's gripper is empty.''
    \item \texttt{slice a/an <sliceable>}: ``the sliceable object is already sliced and the robot's gripper is holding a knife.''
    \item \texttt{put things on the countertops away}: ``all objects on the countertops have been put away in the cabinets and there are no more unexplored countertops or cabinets.''
\end{itemize}

They might not be fully aligned with the human's intention (e.g., someone may think having an object in view already satisfies the goal of ``find'', but the agent believes the task is not done until it picks the object up), but they reflect what the agent would do if the subtask is chosen.
This is very helpful for reusing previously learned tasks.
\subsection{Reproducibility Checklist}

We answer ``partial'' to the following question on the reproducibility checklist
\begin{quote}
    [Computational experiments] If an algorithm depends on randomness, then the method used for setting seeds is described in a way sufficient to allow replication of results.
\end{quote}
This is because our experiments involve the use of GPT-4 and an Unity-based simulator, whose internal mechanism is not fully disclosed to the best of our knowledge.
We have done our best to set the temperature of GPT-4 to $0$, but empirical experiments and experiences from other users suggest that its behavior is still not deterministic when the temperature is set to $0$.
Additionally, there is physics simulation in the simulator (e.g., when slicing a lettuce, the slices will fall apart).
These motions are not deterministic according to our observation, and the documentation of the simulator does not provide a way to make the result deterministic.
Because the bootstrapping process contains multiple steps ($\sim 1400$ steps), small discrepancies at the beginning may accumulate and result in very different production rules learned.

Despite we cannot guarantee whether anyone bootstrapping the agent from scratch will generate the same production rules as we do, we attach (in the code and data supplementary material) our bootstrapped production rules, the logs generated during the bootstrapping process, and the logs generated during the testing process such that one can use them to verify the results we reported in the experiments section of the main paper.

\end{document}